\documentclass[manuscript,screen,nonacm]{acmart}

\usepackage{multirow}
\usepackage{booktabs}
\usepackage{graphics}
\usepackage{adjustbox}
\usepackage{subcaption}
\usepackage[font=small,labelfont=bf]{caption}
\usepackage{url}
\usepackage{graphicx}
\usepackage{algpseudocode}
\usepackage{algorithm}
\usepackage{tabularx}

\AtBeginDocument{%
  }

\setcopyright{acmlicensed}
\copyrightyear{2018}
\acmYear{2018}
\acmDOI{XXXXXXX.XXXXXXX}

\acmConference[Conference acronym 'XX]{Make sure to enter the correct
  conference title from your rights confirmation emai}{June 03--05,
  2018}{Woodstock, NY}
\acmISBN{978-1-4503-XXXX-X/18/06}

\begin{document}

\title{Towards A Comprehensive Visual Saliency Explanation Framework for AI-based Face Recognition Systems}


\author{Yuhang Lu}
\email{yuhang.lu@epfl.ch}
\author{Zewei Xu}
\email{xuzewei@hotmail.com}
\author{Touradj Ebrahimi}
\email{touradj.ebrahimi@epfl.ch}
\affiliation{%
  \institution{EPFL}
  \city{Lausanne}
  \country{Switzerland}
}








\renewcommand{\shortauthors}{Yuhang Lu et al.}

\begin{abstract}
Over recent years, deep convolutional neural networks have significantly advanced the field of face recognition techniques for both verification and identification purposes. Despite the impressive accuracy, these neural networks are often criticized for lacking explainability. There is a growing demand for understanding the decision-making process of AI-based face recognition systems. Some studies have investigated the use of visual saliency maps as explanations, but they have predominantly focused on the specific face verification case. The discussion on more general face recognition scenarios and the corresponding evaluation methodology for these explanations have long been absent in current research. Therefore, this manuscript conceives a comprehensive explanation framework for face recognition tasks. Firstly, an exhaustive definition of visual saliency map-based explanations for AI-based face recognition systems is provided, taking into account the two most common recognition situations individually, i.e., face verification and identification. Secondly, a new model-agnostic explanation method named CorrRISE is proposed to produce saliency maps, which reveal both the similar and dissimilar regions between any given face images. Subsequently, the explanation framework conceives a new evaluation methodology that offers quantitative measurement and comparison of the performance of general visual saliency explanation methods in face recognition. Consequently, extensive experiments are carried out on multiple verification and identification scenarios. The results showcase that CorrRISE generates insightful saliency maps and demonstrates superior performance, particularly in similarity maps in comparison with the state-of-the-art explanation approaches.
\end{abstract}





\maketitle

\section{Introduction}

Recent years have witnessed great advances in face recognition due to the rapid development of deep learning techniques. Current deep learning-based face recognition systems achieve exceptional performance on established benchmarks and have been widely deployed in several applications, including but not limited to access control and surveillance. However, the predictions made by these systems tend to be challenging to interpret. The deployment of such biometric systems poses a potential threat to privacy and data protection, resulting in serious public concern. To address these issues, it is essential to comprehend and explain the behavior of face recognition systems, thereby improving their performance and making them more widely accepted in society. 

Early on, some studies~\cite{zhuang2010facial, kortylewski2019analyzing, terhorst2021comprehensive} have exposed the bias problem of specific deep learning-based face recognition models concerning ethnicity, gender, and age. Lu et al.~\cite{lu2022novel} further enhanced their transparency by investigating the performance in the presence of various realistic influencing factors. However, these studies primarily focus on revealing the weaknesses of face recognition systems, while overlooking the explanation of the decision-making process. Existing deep learning-based face recognition systems often rely on a complicated and unintuitive process to reach a final decision, often referred to as a “black box”. The lack of interpretation of these decisions can undermine user trust and hinder the governance of face recognition technology. 

In recent years, visual saliency algorithms have emerged as the prevailing approach to explain AI-based decision models in vision tasks, highlighting internal neural network layers~\cite{zeiler2014visualizing, olah2017feature, bau2017network} or crucial pixels in the input image that are relevant to the model’s decision~\cite{simonyan2013deep,binder2016layer,zhou2016learning,zhang2018top,selvaraju2017grad,chattopadhay2018grad,li2018tell}. While these algorithms have achieved impressive results, they predominantly address explanation problems in image classification and detection tasks. In this work, we focus on the crucial problem of explainable face recognition, specifically by developing insightful explainability tools to interpret the decision-making process of a deep learning-based face recognition system. 

The face recognition task mainly comprises two scenarios, i.e., \textit{face verification} and \textit{face identification}~\cite{wang2021deep}. The former determines whether one face image matches with another, while the latter aims to identify the subject of a given face image from an entire face database. The given face images and those from the database are called \textit{probe} and \textit{gallery} images. In general, face recognition differs from other vision tasks not only due to the notable difference in the output formats but also the decision-making process, which often involves two (face verification) or more (face identification) input face images. While existing studies~\cite{castanon2018visualizing, williford2020explainable, lin2021xcos, mery2022true, knoche2023explainable, huber2024efficient, lu2024towards, lu2024explainable} have devised various saliency algorithms to reveal important pixels, explaining a face recognition model entails more than merely highlighting the critical areas using saliency maps. Beyond this, it should also interpret why given face images are perceived as similar or dissimilar to the recognition system, aiding in the subsequent decision-making process. Furthermore, current research has predominantly concentrated on explaining face verification tasks, neglecting another equally important scenario of face identification. In this context, this manuscript contributes a comprehensive explanation framework for AI-based face recognition systems. 

Firstly, we refine the prevailing definition of visual saliency map-based explanations tailored for learning-based face recognition systems. Specifically, an explainable face verification system should reveal similar regions when the model determines the input pair of images as ``matching'', and conversely, the dissimilar regions when it gives a ``non-matching'' decision. Comparably, an explainable face identification system should elucidate the similarities between the probe image and top-ranking gallery images. 
To our acknowledgment, this is the first work to comprehensively investigate both face verification and identification scenarios. 
Then, a Correlation-based Randomized Input Sampling for Explanation (CorrRISE) algorithm is proposed. It is model-agnostic and capable of providing saliency maps that adhere to the aforementioned explainable face recognition definition and highlight similar and dissimilar regions between any input face images. Moreover, this manuscript proposes a new objective evaluation methodology to quantitatively compare different state-of-the-art explainable face recognition methods. 

This manuscript is an extended version of our recent publication~\cite{lu2024towards} and introduces a comprehensive saliency-based explainable face recognition framework. It broadens the commonly studied explainable face verification problem to more generic scenarios and intends to address the more exhaustive and challenging problem of explainable face recognition, provided with conceptual, technical, and experimental updates. In summary, the following contributions have been made:
\begin{itemize}
\item This manuscript provides a comprehensive definition of the explainable face recognition problem, taking into account the two most practical recognition scenarios
\item A novel model-agnostic explanation method called CorrRISE is proposed, which highlights the similarity and dissimilarity regions between any given face images
\item A new evaluation methodology is conceived to quantitatively measure the performance of general saliency map-based explanation methods for face recognition
\item Substantial experiments on multiple face verification and identification scenarios have been carried out and presented, followed by a detailed quantitative comparison with the current state-of-the-art explanation methods, demonstrating the effectiveness of the proposed method
\end{itemize}

\section{Related Work}
\label{relatedwork}

\subsection{Visual Explanation via Saliency Maps}
In the field of explainable artificial intelligence (XAI), visual saliency algorithms have been widely used to explain decision systems in vision tasks that rely on deep learning techniques. A saliency map is essentially an image where each pixel value represents the importance of the corresponding pixel. This map helps identify the significant areas of an input image that contribute to the final output of a ``black-box'' model. In general, there are two types of methods to create such saliency maps. 

The first category of methods involves backpropagating an importance score from the model's output to the input pixels through the neural network layers. Notable examples includes Gradient Backpropagation~\cite{simonyan2013deep}, Layer-wise Relevance Propagation~\cite{binder2016layer}, Class Activation Maps (CAM)~\cite{zhou2016learning}, and Excitation Backpropagation~\cite{zhang2018top}. Many of these methods require access to the intrinsic architecture or gradient information of the model. 
Grad-CAM~\cite{selvaraju2016grad} and Grad-CAM++~\cite{chattopadhay2018grad} generalized CAM to be applied to arbitrary convolutional neural networks (CNNs) by weighing the feature activation values with the class-specific gradient information that flows into the final convolution layer. Considering their fame and flexibility in CNNs-based models, this manuscript adapts them to the explainable face recognition problem to compare with our proposed method. 

The second category of methods performs random perturbations on the input image, such as adding noise or occlusion, and produces saliency maps by observing the impact on the model's output~\cite{zeiler2014visualizing, ribeiro2016should, fong2017interpretable, dabkowski2017real, petsiuk2018rise}. For example, Zeiler et al.~\cite{zeiler2014visualizing} masked square parts of an image with a sliding window and determined the importance by observing the drop in classification accuracy. Ribeiro et al.~\cite{ribeiro2016should} proposed an interpretable approximate linear decision model (LIME) in the vicinity of a particular input, which analyzes the relation between the input data and the prediction through a perturbation-based forward propagation. The RISE~\cite{petsiuk2018rise} algorithm generates random masks, applies them to the input image, and utilizes the output class probabilities as weights to compute a weighted sum of the masks as a saliency map.

\subsection{Explainable Face Recognition}
While most XAI techniques involving saliency are developed for image classification, there is a growing demand for explanation methods in other image understanding tasks, such as object detection~\cite{petsiuk2021black} and image similarity search and retrieval~\cite{stylianou2019visualizing, dong2019explainability, hu2022x}.
In face recognition, earlier endeavors~\cite{castanon2018visualizing, williford2020explainable} primarily adapted saliency-based explanation algorithms~\cite{zhang2018top, petsiuk2018rise, selvaraju2016grad, selvaraju2017grad} from classification tasks. Alternative research directions focus on face verification models that are explainable by themselves, often referred to as intrinsic explanation methods. For example, Yin et al.~\cite{yin2019towards} designed a feature activation diverse loss to encourage learning more interpretable face representations. Lin et al.~\cite{lin2021xcos} proposed a learnable module that can be integrated into face recognition models and generate meaningful attention maps. Xu et al.~\cite{xu2023discriminative} leveraged a face reconstruction module to localize discriminative face regions. However, these self-explained models need to be trained exclusively and thus are impractical for third-party deployed recognition systems. 
Instead, recent studies offer more flexible solutions by leveraging gradient backpropagation~\cite{huber2024efficient, lu2024explainable}, which do not need to access or retrain the face recognition model. Another category of methods provides purely ``black-box'' explanations for arbitrary face recognition models. Mery et al~\cite{mery2022true, mery2022black} introduced several perturbation-based methods to create explainable saliency maps without altering or retraining the model, yielding visually promising results. xFace~\cite{knoche2023explainable} further improved them by applying more systematic occlusions to inputs and measuring the feature distance deviations.

However, all the prior studies have been designed solely to explain the face verification process but neglect the other common scenario of face identification. To bridge this gap, this manuscript introduces a comprehensive explanation framework, encompassing detailed definitions for both explainability scenarios. Furthermore, we adapt the existing state-of-the-art explanation techniques to accommodate the identification scenario for a more straightforward comparison.

\subsection{Objective Evaluation Methodology for Explainable Face Recognition}
Despite the advancement of various explanation algorithms in vision tasks, the evaluation of these approaches has primarily relied on visualizations, making it difficult to compare them. In particular, only a few metrics have been developed for assessing visual saliency explanation tools. In the past, human evaluation was the predominant way to evaluate model explainability~\cite{herman2017promise, zhang2018top}. For example, Zhang et al.~\cite{zhang2018top} carried out a ``pointing game'' in an explainable classification task, which counts the number of saliency points contained within a human-annotated bounding box of an object. Petsiuk et al.~\cite{petsiuk2018rise} proposed two automatic objective metrics, tailored for the image classification task. These metrics measure the change in output classification probability upon removal or addition of salient pixels from or to the input image.

In face verification, Williford et al.~\cite{williford2020explainable} played an ``inpainting game'' that leveraged a triplet of images comprising a probe image, a matching image, and an inpainted image of the same subject. It examined whether the explanation method can correctly localize the discriminative pixels within the inpainted regions while excluding other identical pixels. However, constructing such an inpainting dataset requires substantial effort, involving the manipulation of thousands of images. Castanon et al.~\cite{castanon2018visualizing} quantified the visualized discriminative features through a ``hiding game'' on a standard face verification dataset and protocol. This approach iteratively obscures the least important pixels in the image, ranked according to the saliency map. However, it is not precise enough to differentiate high-performing explanation methods according to the assessment results reported in~\cite{xu2023discriminative}.

\begin{figure*}
     \centering
     \begin{subfigure}[b]{0.48\textwidth}
         \centering
         \includegraphics[width=\textwidth]{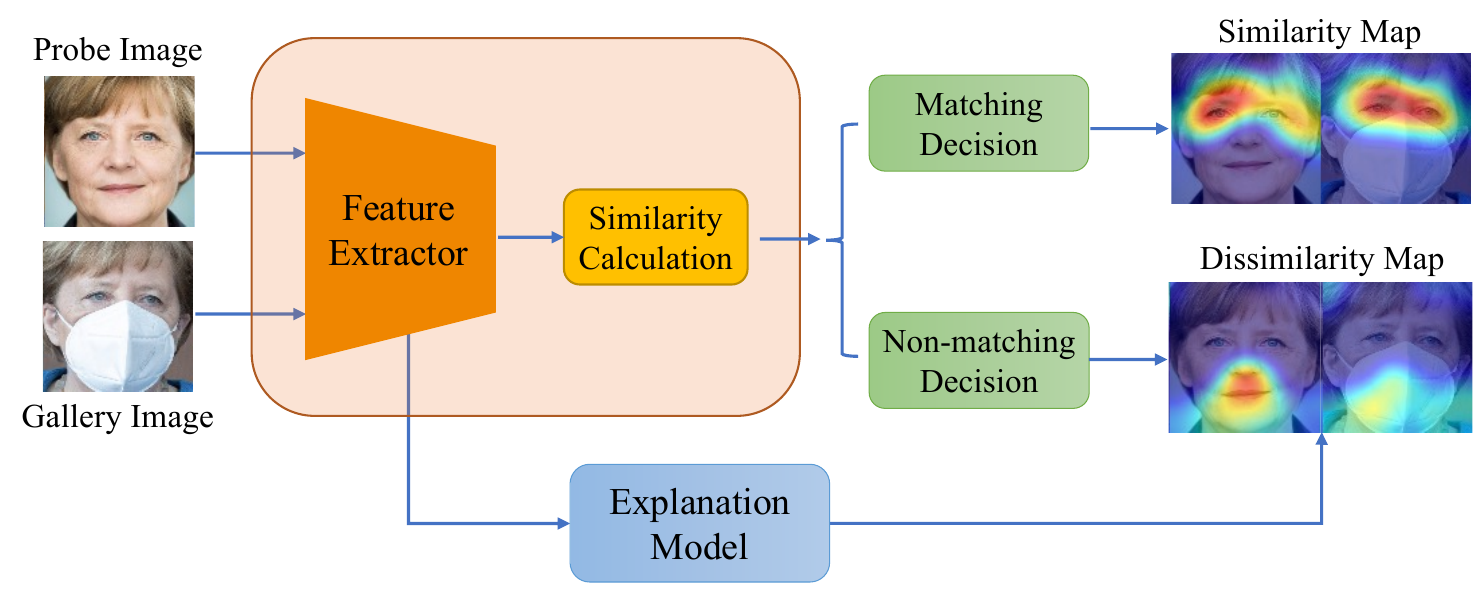}
         \caption{Saliency explanation for face verification.}
         \label{fig:verDefinition}
     \end{subfigure}
     \hfill
     \begin{subfigure}[b]{0.48\textwidth}
         \centering
         \includegraphics[width=\textwidth]{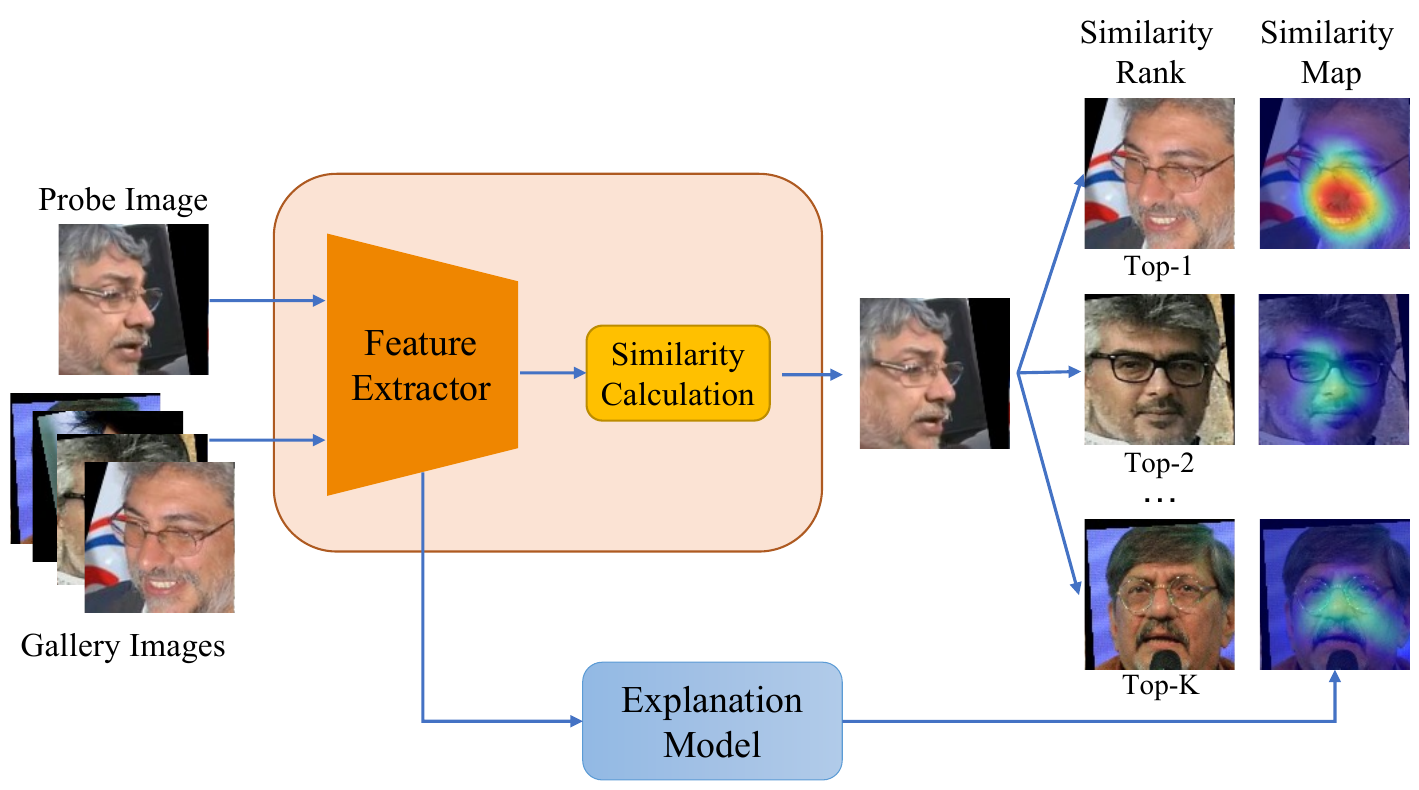}
         \caption{Saliency explanation for face identification.}
         \label{fig:idenDefinition}
     \end{subfigure}
     \caption{The proposed definition of visual saliency-based explanations in two typical face recognition scenarios, i.e., face verification and identification.}
\end{figure*}

\section{Proposed Method}
\label{method}

\subsection{Definition of Explanation for Face Recognition}
\label{definition}
A face recognition system gives different types of decisions depending on whether it is employed in verification and identification scenarios. In face verification, the system computes the cosine similarity score between two input face images and gives a ``matching'' decision if the score surpasses a predetermined threshold, otherwise, it makes a ``non-matching'' decision. In face identification, the system calculates a one-to-many similarity matrix between a probe face and numerous gallery faces from the database and then determines the specific identity of the probe image based on the most similar gallery image. Thus, this manuscript provides two distinct definitions for explainable face verification and identification.

\subsubsection{Definition for Explainable Face Verification}
An earlier study~\cite{castanon2018visualizing} approached the explainability of a face verification model similarly to that of an image classification model, aiming to visualize the discriminative information within each individual face image. However, the verification process typically involves two images. The global critical regions of each image identified by the model may not inherently be similar or dissimilar areas between two images. 
Williford et al.~\cite{williford2020explainable} utilized a face triplet, i.e., probe, matching gallery, and non-matching gallery, to explain the relative importance of facial regions. They defined explainable face verification as a way to highlight specific regions of the probe image which simultaneously maximize the similarity with the matching image and minimize the similarity with the non-matching. 
However, the face verification process operates independently for a pair of inputs instead of a triplet, making this definition less practical.
Mery~\cite{mery2022true} improved the definition by directly exploring the relevant parts between two images when a match is established. Nevertheless, their definition overlooks the irrelevant parts between inputs, which particularly dominate the decision-making process for non-matching pairs of images.

As shown in Fig.~\ref{fig:verDefinition}, this manuscript defines the problem of explainable face verification as follows. 
Given a pair of images feeding into a face verification system, the explanation method should generate corresponding saliency maps for both input images, which should clearly interpret the prediction results by answering the following questions:

\begin{itemize}
    \item If the face verification system believes the input pair is matching, which regions are \textbf{similar} to the model?
    \item If the face verification system believes the input pair is non-matching, which regions are \textbf{dissimilar} to the model?
\end{itemize}

\subsubsection{Definition for Explainable Face Identification}
Previous research has proactively explored the problem of explainable face verification, whereas, to the best of our knowledge, there is no existing work that interprets the face identification process using saliency maps. 
Fig.~\ref{fig:idenDefinition} shows a typical process of face identification, where a face identification model computes the 1:n similarity and then determines the subject of the given probe image based on the most similar face from the gallery database. In practice, the correct subject for the probe image often appears in the top-K most similar gallery images, depending on the efficacy of the model or the difficulty of identifying the specific subject. Therefore, an explanation method should help understand why the identification model ranks a subject over others.

As illustrated in Fig.~\ref{fig:idenDefinition}, this work outlines the definition of explainable face identification as follows: Given a probe face image and numerous gallery images, the explanation method should generate saliency maps for the top-K gallery images, highlighting the pixels that are similar to the probe image respectively. In principle, K can be selected to any number larger than one based on the need for explainability. For demonstration purposes, K $=5$ is selected in this manuscript. 

\begin{figure*}[t]
	\centering
	\begin{adjustbox}{width=\textwidth}
    \includegraphics[]{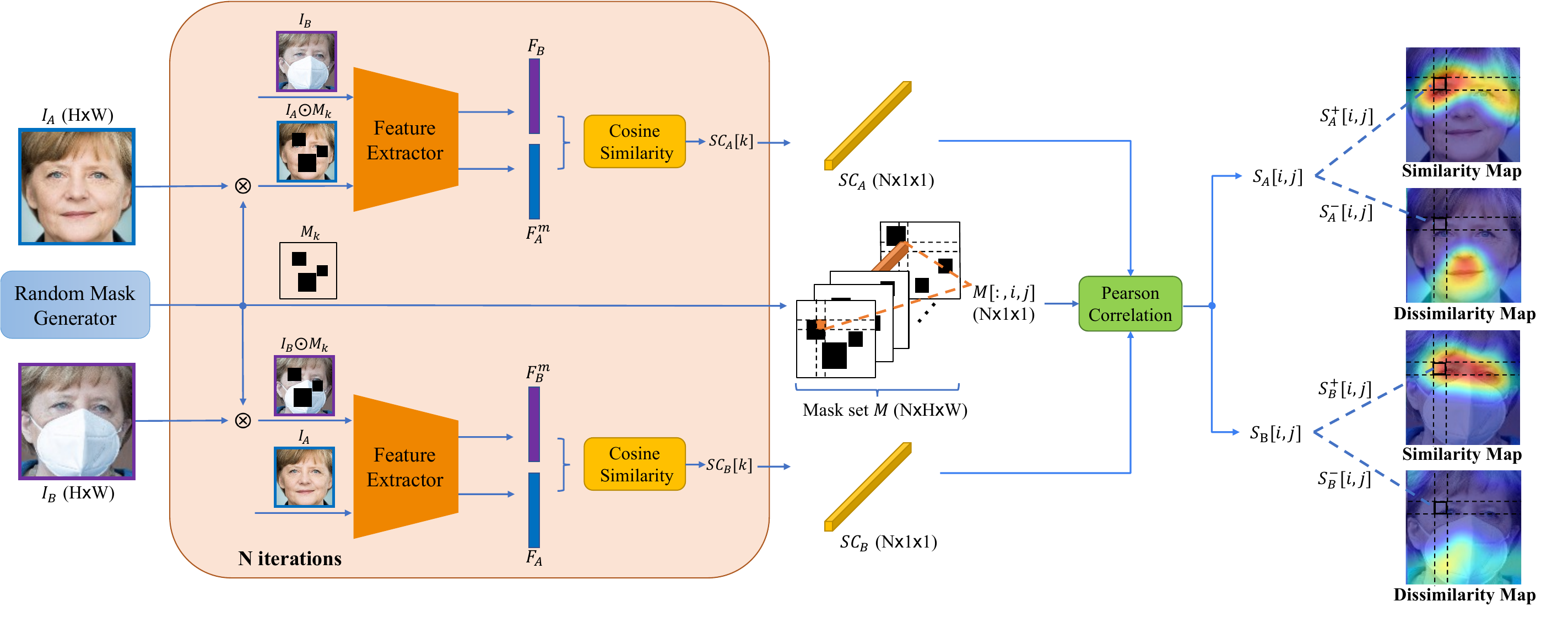}
	\end{adjustbox}
	\caption{Workflow of the proposed CorrRISE explanation method. The similarity and dissimilarity maps are calculated respectively given an arbitrary input face pair. The block in the middle repeats $N$ iterations using different random masks. The output similarity scores and the mask set are fed to the correlation module to calculate similarity and dissimilarity saliency maps in a pixel-wise manner. }
	\label{fig:corrrise}
\end{figure*}

\subsection{Proposed explanation method}

This section presents a new model-agnostic explanation method called CorrRISE to address the new definition of explainable face recognition, see Fig.~\ref{fig:corrrise}. 
In principle, CorrRISE generates saliency maps by injecting perturbation and observing the impact on output. Thus, it provides ``black-box'' explanations and can be applied to any face recognition system without retraining or access to the network. In contrast with other perturbation-based approaches explaining classification models, CorrRISE applies random masks to face images and measures the effect of masked regions on the final similarity scores between two faces rather than a single categorical output. Subsequently, the Pearson correlation between a list of similarity scores and random masks is calculated in a pixel-wise manner to obtain saliency maps. The similar and dissimilar pixels are disentangled from the saliency map according to the correlation coefficients. This innovative approach distinguishes CorrRISE from previous perturbation-based explainable face recognition methods~\cite{mery2022true, knoche2023explainable} and prior RISE adaptations proposed by~\cite{williford2020explainable,mery2022black}.

As illustrated in Fig.~\ref{fig:corrrise}, CorrRISE operates in a pair-wise manner and produces saliency maps for any given two input images, which naturally aligns our definition for explainable face verification. Whilst, the identification process calculates 1:n similarity, which can be practically represented as a repetition of the 1:1 verification process. Therefore, the CorrRISE explanation method can be directly applied to generate saliency maps for the face identification scenario without modifying the method itself. 

The CorrRISE algorithm comprises two pivot steps, i.e., mask generation and saliency map generation. The detailed procedures are elaborated as follows.

\begin{algorithm}[t]
\caption{}
\label{alg:corrrise}
\begin{algorithmic}
\Procedure{CorrRISE}{} \\
\hspace*{\algorithmicindent}\textbf{Input:} number of iterations $N$, face recognition model $f_x$, similarity function $\texttt{Score()}$, face images $I_A$ and $I_B$\\
\hspace*{\algorithmicindent}\textbf{Output:} saliency maps $S^+_A, S^-_A, S^+_B, S^-_B$ 

\State $H, W \gets$ \texttt{Size($I_A$)}
\For{$k=1:N$}
\State $M_k \gets \texttt{RandomMaskGenerator}(H,W)$
\State $x_A, x_B \gets f_x(I_A), f_x(I_B)$
\State $x_A^m, x_B^m \gets f_x(I_A \odot M_k), f_x(I_B \odot M_k)$
\State $SC_{A}[k] \gets \texttt{Score}(x_A^m, x_B)$
\State $SC_{B}[k] \gets \texttt{Score}(x_B^m, x_A)$
\State $M[k,:,:] \gets M_k$
\EndFor

\For{$i=1:H$}
\For{$j=1:W$} \\
\hspace*{\algorithmicindent}\hspace*{\algorithmicindent}\hspace*{\algorithmicindent} $S_A[i,j] \gets $ \texttt{PearsonCorr}($SC_A, M[:,i,j]$) 
\\
\hspace*{\algorithmicindent}\hspace*{\algorithmicindent}\hspace*{\algorithmicindent} $S_B[i,j] \gets $ \texttt{PearsonCorr}($SC_B, M[:,i,j]$)

\EndFor
\EndFor

\State $S^+_A, S^-_A \gets S_A[S_A\ge0], S_A[S_A<0]$
\State $S^+_B, S^-_B \gets S_B[S_B\ge0], S_B[S_B<0]$

\EndProcedure
\end{algorithmic}
\end{algorithm}

\subsubsection{Mask Generation}

Mask generation is an essential step that injects random perturbations into the input. The random mask generator in Fig.~\ref{fig:corrrise} randomly samples multiple small square patches in various locations on a plain image. As illustrated in the figure, the values of patches are set to 0 and all the patches constitute a binary mask, which occludes the corresponding pixels in a face image. We additionally test a variety of approaches to generate the patch values, such as purely random initiation, bilinear interpolation between [0, 1], and Gaussian distribution, while the binary mask obtains the best performance. 
In summary, the mask generation steps are as follows.
\begin{enumerate}
    \item Initialize the parameters of the mask generator, i.e., the total number of masks $N$, and the number and size of square patches in each mask.
    \item Sample multiple square patches with zero values in random locations of each mask $M_i$ and finally get the mask set $\{M_i, i=1,...,N\}$.
    \item Inject perturbation by multiplying the mask and input images.
\end{enumerate}

\subsubsection{Correlation-based Saliency Map Generation}

Fig.~\ref{fig:corrrise} illustrates an overview of the proposed CorrRISE method and Algorithm~\ref{alg:corrrise} presents detailed steps for saliency map generation. In general, given a pair of images $\{I_A, I_B\}$ and a face recognition model $f_x$, the objective is to produce saliency maps highlighting both the similar and dissimilar regions between two faces. First, as shown in the previous step, CorrRISE leverages a mask generator to randomly produce $N$ masks $M=\{M_i, i=1,...,N\}$. Each mask $M_i$ is then multiplied with the corresponding input image, e.g., $I_A$. The masked $I_A\odot M_i$ and unmasked $I_B$ are fed into the face recognition model $f_x$ to capture the deep face representation $\{x^m_A, x_B\}$. The cosine similarity score $SC_i$ between the deep features is then calculated. After iterating all the $N$ masks, the list of scores $SC=\{SC_i, i=1,...,N\}$ corresponding to the mask list is recorded. Subsequently, Pearson correlation is performed between $SC$ and $M$ on a pixel-wise basis to obtain the final saliency map $S_A$ for $I_A$. The location of positive correlation coefficients represents the regions on $I_A$ that are similar to $I_B$, while the locations of negative coefficients represent the dissimilar regions. The same procedure is replicated for $I_B$ to obtain the saliency map $S_B$. 

The generation of two saliency maps is conducted separately for each image, as depicted in Fig.~\ref{fig:corrrise}. Because a face recognition system can mistakenly match two irrelevant but both masked face images, interfering with the computation of similarity scores during generation. Furthermore, while the saliency map generation of CorrRISE is grounded in the face verification process, it can be easily adapted to the identification scenario by replicating the generation process for the same probe image and various gallery images.

\begin{figure}[t]
	\centering
    \includegraphics[width=0.8\linewidth]{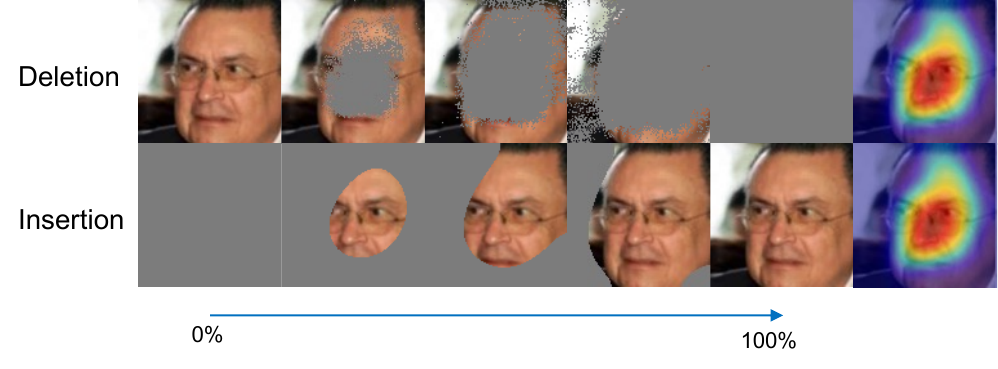}
	\caption{The deletion and insertion processes to calculate corresponding evaluation metrics. The most important pixels indicated by the saliency maps are gradually removed/added.}
	\label{fig:del_ins}
\end{figure}

\section{Evaluation Methodology}
\label{chapter:eval}
Despite the promising development of explainability methods in vision tasks, the importance of rigorous objective evaluation methodologies has long been overlooked. In particular, only a few objective metrics or protocols have been designed for visual saliency explanation tools. The evaluation of previous explainable face recognition methods has been mostly based on visualization, making it difficult to compare to others.

\begin{algorithm}[t]
\caption{Deletion and Insertion Metric for Face Verification Scenario}
\label{alg:metricsVer}
\begin{algorithmic}
\Procedure{Evaluation Metric}{} \\
\hspace*{\algorithmicindent}\textbf{Input:} face recognition model $f_x$, evaluator $\texttt{eval()}$, testing dataset $D$, saliency maps $S$, number of steps $n$\\
\hspace*{\algorithmicindent}\textbf{Output:} deletion score $d_1$, insertion score $d_2$
\State $S \gets \texttt{GaussianBlur}(S)$ 
\State $S \gets \texttt{Sorting}(S)$ \algorithmiccomment{sort in descending order}
\For{$i=1:n$}
\State $p \gets i/n\times100$
\State Mask the first $p\%$ pixels of $D$ to get $D^\prime_{1}$
\State Insert the first $p\%$ pixels to plain images to get $D^\prime_{2}$
\State $acc_{1i} \gets \texttt{eval}(f_x, D^\prime_{1})$
\State $acc_{2i} \gets \texttt{eval}(f_x, D^{\prime}_{2})$
\EndFor
\State $d_1 \gets \texttt{AreaUnderCurve}(acc_{1i}$ vs. $i/n,\text{ }i=1:n)$
\State $d_2 \gets \texttt{AreaUnderCurve}(acc_{2i}$ vs. $i/n,\text{ }i=1:n)$
\EndProcedure
\end{algorithmic}
\end{algorithm}

This manuscript contributes new ``Deletion'' and ``Insertion'' metrics to better assess explanation methods for the general face recognition task, including both verification and identification scenarios. In principle, these metrics measure the change in the recognition performance after modifying the input image according to the importance map generated by the explanation method. The intuition behind the proposed metrics is that an effective saliency map is expected to precisely highlight the most important regions of two faces with the smallest number of pixels, based on which the face recognition model makes final decisions. The faster the overall recognition performance drops/rises after removing/adding salient pixels, the more accurate the produced saliency map. Changes in recognition performance are measured by two separate auxiliary face verification and identification tasks, aligning with the respective face recognition scenarios.

Algorithm~\ref{alg:metricsVer} elaborates on the detailed procedures for calculating the proposed metrics in the face verification scenario. The ``Deletion'' metric executes the following steps. First, the generated saliency map for each input image is sorted according to the importance value. Then, $n$ threshold values are evenly sampled from $[0,1]$, i.e., $p_k = k/n\times 100\%$ where $k\in \{1, \dots, n\}$, and serve as different percentages of modified pixels in the input image. Subsequently, a verification task using dataset $D$ is performed iteratively as an auxiliary task in the overall evaluation methodology. In each iteration, $p_k$ percent of the most salient pixels are masked from the input image of the entire testing dataset, as shown in Fig.~\ref{fig:del_ins}. The accuracy of the face recognition model is then measured on the modified dataset. Finally, the ``Deletion'' metric is defined as the Area Under the Curve (AUC) score of the Accuracy vs. Percentage of masked pixels curve. The lower the metric, the more accurate the evaluated saliency map. The ``Insertion'' metric takes a complementary approach but follows similar procedures. During each iteration, $p_k$ percent of the most important pixels are inserted into a plain image. Ideally, a higher ``Insertion'' score corresponds to a better explanation of the saliency map.
Practically, similarity and dissimilarity maps are evaluated separately using matching and non-matching parts of the testing dataset.

In the face identification scenario, we take standard identification evaluation as an auxiliary task and quantify the changes in Rank-N identification rate as ``Deletion'' and ``Insertion'' metrics. While saliency maps of gallery images are presented as explanations in our definition, it is impractical to directly apply similar Deletion and Insertion processes to them for evaluation. Because each probe image corresponds to numerous gallery images, which makes the evaluation process considerably inefficient. Instead, we tend to evaluate the saliency maps of their probe counterparts. Specifically, for deletion metric, we first produce K pairs of saliency maps between the probe and its top-K similar gallery images and take an average of those saliency maps belonging to the probe image. Then, a deletion process is applied to the probe image by removing a certain amount of pixels from it and subsequently identifying the correct subject out of the gallery images. The Rank-N score is used in the auxiliary identification task to measure the accuracy of the produced saliency map. Intuitively, the lower the identification score, the more important the obscured regions, and thus the more accurate saliency maps. 
Overall, the “Deletion” metric is defined as the AUC score of the Rank-N vs. Percentage of masked pixels curve, and the lower the better, while the “Insertion” metric takes a complementary approach but follows similar procedures, the higher the better.

\section{Experimental Results}
\label{results}

\subsection{Implementation Details}
\subsubsection{Explanation Method Setup}
\label{label:implementation}
The proposed CorrRISE explanation method does not require any training or access to the internal architecture of the face recognition model. During the explanation process, the default number of generated masks, i.e., the number of iterations, is set to 1000. For each mask, there are 10 patches and the size of each patch is 30$\times$30 pixels.

\subsubsection{Face Recognition Model Setup}
Extensive experiments are conducted using the popular ArcFace~\cite{deng2019arcface} model with ResNet-50~\cite{he2016deep} backbone. 
To demonstrate the generalization capability of our proposed explanation method across various face recognition models, its explainability performance is additionally tested on two face recognition models employing distinct loss functions, i.e., AdaFace~\cite{kim2022adaface} and MagFace~\cite{meng2021magface}. All the face recognition models are trained on the same dataset by running their official publicly available codes.

\subsubsection{Dataset}
The face recognition models are trained with the MS1M~\cite{guo2016ms-celeb-1m} dataset cleaned by Deng et al.~\cite{deng2019arcface}. For evaluation, this manuscript first selects samples from various datasets, namely LFW~\cite{huang2008labeled}, CPLFW~\cite{zheng2018cross}, LFR~\cite{elharrouss2020lfr}, Webface-Occ~\cite{huang2021face}, and IJB-C~\cite{maze2018iarpa}, for visual demonstration in a variety of recognition scenarios. In the proposed objective evaluation methodology, LFW, CPLFW, and CALFW datasets are employed for quantitative evaluation in the verification scenario, while the IJB-C dataset is utilized for the identification scenario. 
All the images are cropped and resized to 112x112 pixels.


\begin{figure}[t]
	\centering
	\begin{adjustbox}{width=0.5\linewidth}
    \includegraphics[]{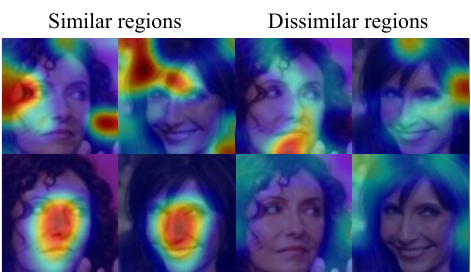}
	\end{adjustbox}
\caption{Sanity check for the CorrRISE explanation method. The first row is the explanation heatmap for a deep model with randomized parameters, while the second row is for a normal face recognition model. The importance increases from blue to red color.}
\label{fig:sanity}
\end{figure}

\subsubsection{State-of-the-art Explanation Methods Setup}
For comparison, several state-of-the-art explainable face recognition methods, namely MinPlus~\cite{mery2022true}, xFace~\cite{knoche2023explainable}, xSSAB~\cite{huber2024efficient}, and FGGB~\cite{lu2024explainable}, have been tested based on their official open-source code. 
MinPlus and xFace methods offer several variations in their publications and the best-performing ones (``AVG'' for MinPlus and ``Method-1'' for xFace) are selected. 
Notably, although all these methods were originally designed for the face verification task, we adapt their code to the identification scenario following the definition in Section~\ref{definition} and further accelerate them by performing the computation in batches on GPU. 
Additinally, several XAI methods~\cite{selvaraju2017grad,chattopadhay2018grad,ribeiro2016should,petsiuk2018rise} have been adapted and tested. For Grad-CAM~\cite{selvaraju2017grad} and Grad-CAM++~\cite{chattopadhay2018grad}, instead of backpropagating the gradients of class-wise posterior probability to activation layers, we adapt them by performing backpropagation for gradients of similarity scores between two input images. Moreover, the third-party adaptation from authors of~\cite{mery2022true} is utilized for LIME~\cite{ribeiro2016should} and RISE~\cite{petsiuk2018rise}.

\subsection{Sanity Check}

A recent study~\cite{adebayo2018sanity} has raised doubts about the reliability of visual saliency methods, suggesting that the produced explanation heatmaps can be independent of the deep model or the input data. 
To address this concern, they introduced a model parameter randomization test for a sanity check. 
In the context of face recognition, an explanation method may provide visually compelling heatmaps by directly emphasizing the center of the faces without interacting with the face recognition model. Therefore, this manuscript employs a similar sanity check to validate the effectiveness of the proposed method. Specifically, the parameters of the ResNet-50 backbone network are randomly initialized and then the CorrRISE algorithm is applied to the randomized model. 
The first row of Fig.~\ref{fig:sanity} shows that the CorrRISE algorithm will generate nonsensical saliency maps when attempting to explain a face recognition model with fake parameters. This result indicates that the proposed explanation method relies on a valid recognition model and is capable of producing meaningful interpretations.



\begin{figure*}[t]
	\centering
	\begin{adjustbox}{width=0.9\textwidth}
    \includegraphics[]{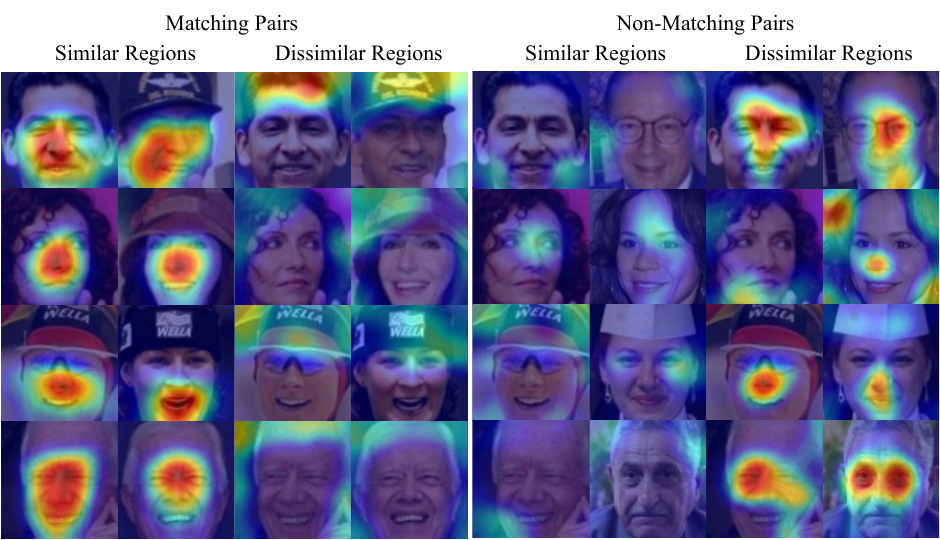}
	\end{adjustbox}
\caption{Visual explanation results from CorrRISE for both matching and non-matching face pairs in standard face verification scenario. The produced saliency maps explain why the verification model makes correct predictions on all face pairs. The saliency value increases from blue to red color. }
\label{fig:verification}
\end{figure*}

\subsection{Visual Explanation Results}

This section presents the visual results of the saliency maps generated by our proposed CorrRISE algorithm. For clarity and fair comparison, all the experiments here are conducted on the ArcFace model. First, the explanation ability of CorrRISE is tested on the standard face verification scenario with face images sampled from LFW and WebFace-Occ datasets. 
Then, the behavior of the face recognition model in several challenging verification scenarios is analyzed and explained through CorrRISE-produced saliency maps, including some failure cases due to very similar subjects or significant head pose variations. Subsequently, CorrRISE is employed in a standard face identification scenario with two illustrative examples presented. 
Finally, a visual comparison with other explanation approaches is provided.


\begin{figure}[t]
	\centering
	\begin{adjustbox}{width=0.5\linewidth}
    \includegraphics[]{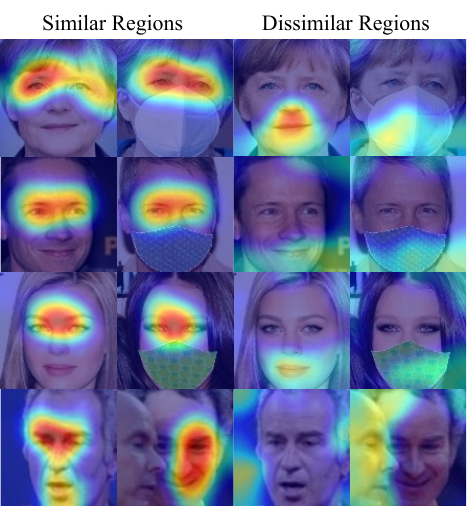}
	\end{adjustbox}
\caption{Saliency map explanations for the predictions of the face recognition model on partially-occluded faces. The masked regions are accurately identified as dissimilar regions. The saliency value increases from blue to red color.}
\label{fig:occlusion}
\end{figure}


\begin{figure}
     \centering
     \begin{subfigure}[b]{0.5\textwidth}
         \centering
         \includegraphics[width=\textwidth]{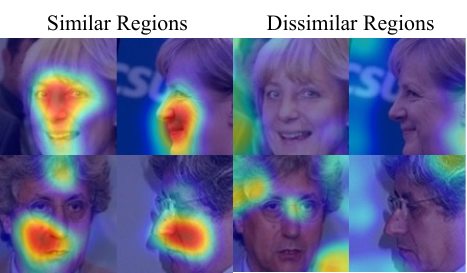}
         \caption{Examples that face recognition model makes correct predictions.}
         \label{fig:pose1}
     \end{subfigure}
     \hfill
     \begin{subfigure}[b]{0.5\textwidth}
         \centering
         \includegraphics[width=\textwidth]{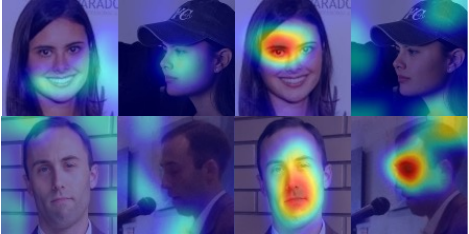}
         \caption{Examples that face recognition model makes wrong predictions.}
         \label{fig:pose2}
     \end{subfigure}
     \caption{Saliency map explanations for the predictions of the face recognition model on four matching but ill-posed pairs of faces. The saliency value increases from blue to red color.}
     \label{fig:pose}
\end{figure}



\begin{figure*}[t]
	\centering
	\begin{adjustbox}{width=0.8\textwidth}
    \includegraphics[]{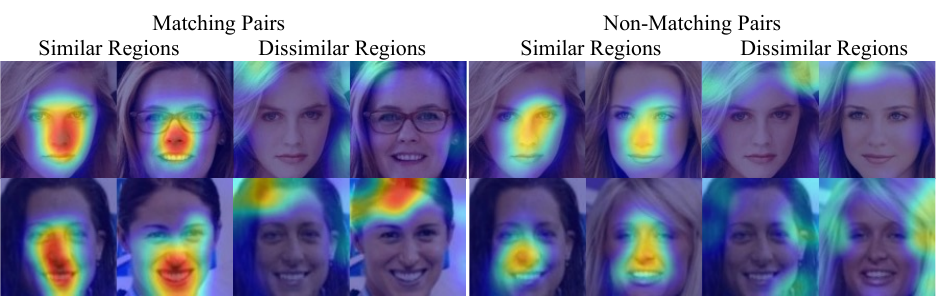}
	\end{adjustbox}
\caption{Saliency map explanation that interprets why the face recognition model mistakenly matches the two non-matching pairs (right). The saliency value increases from blue to red color.}
\label{fig:fail}
\end{figure*}



\begin{figure*}[t]
	\centering
	\begin{adjustbox}{width=0.95\linewidth}
    \includegraphics[]{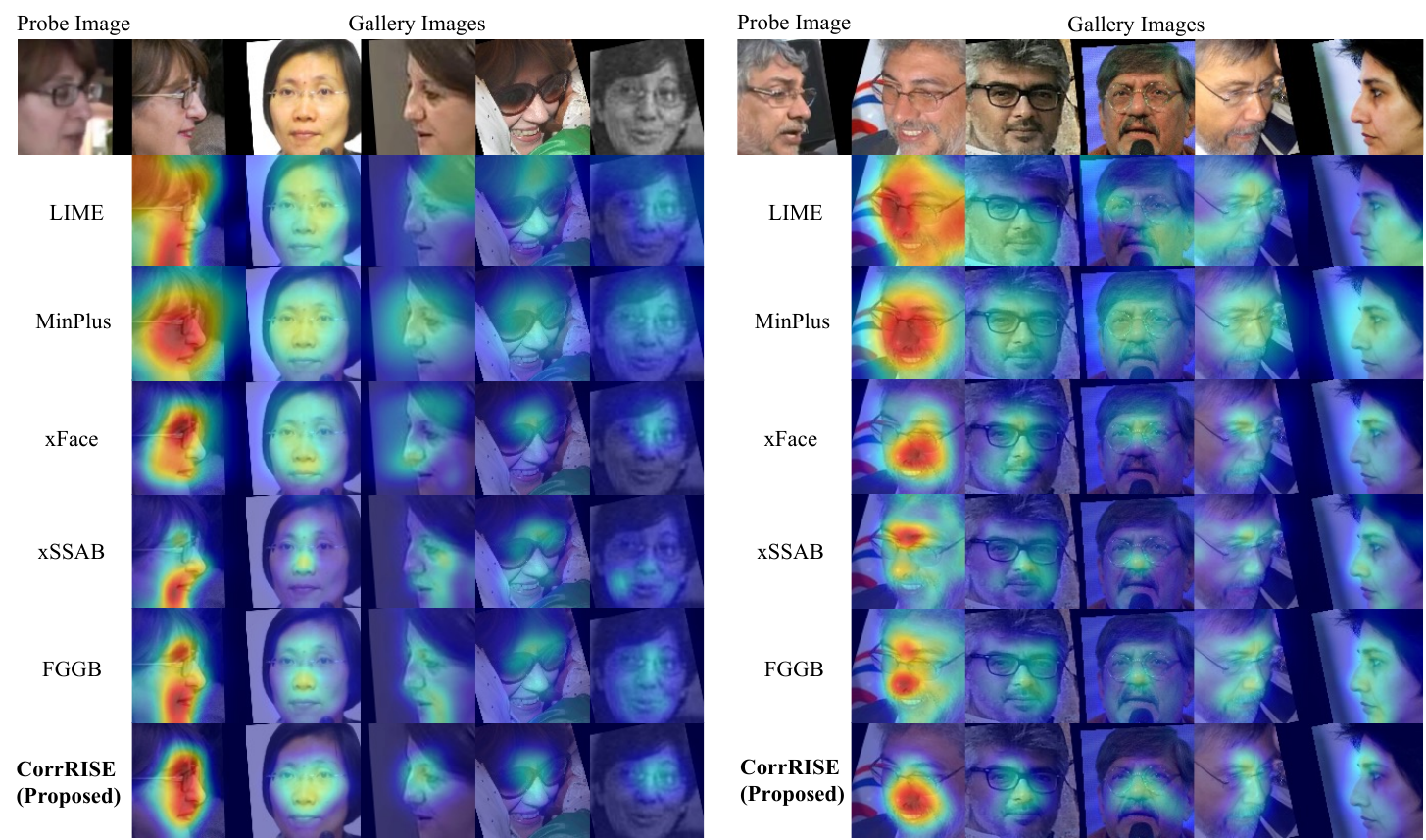}
	\end{adjustbox}
\caption{Two examples produced by CorrRISE and various explanation methods in the explainable face identification scenario. Each example comprises one probe image and the five most similar gallery images selected by the face recognition model. The visual saliency maps are provided for all the gallery images to interpret the model's decision. The importance increases from blue to red color.}
\label{fig:idresults}
\end{figure*}

\subsubsection{Standard Verification Scenario}

Fig.~\ref{fig:verification} illustrates the visual explanations for the model's decision regarding four matching and four non-matching pairs of images taken from the LFW dataset. Remarkably, here the deep model makes correct predictions on all eight pairs.

As a result, the saliency maps produced by CorrRISE properly highlight the similar regions between the matching pairs. Generally, the salient region focuses on eyes, noses, and mouths, while there are variations from person to person. For example, the face recognition model relies more on the cheek when comparing the matching pair in the first row, while it emphasizes the open mouth for the pair in the third row. It is also noteworthy that the dissimilar regions often concentrate on irrelevant backgrounds and unexpected occlusions, such as hats. For the non-matching pairs, CorrRISE attempts to localize the similar areas between the non-matching faces but with relatively low salient values. In contrast, it produces saliency maps that clearly highlight the most dissimilar regions in their faces, indicating very low similarity between them. 

To further show the effectiveness of the proposed explanation algorithm, an additional test has been conducted using partially occluded faces from the WebFace-Occ dataset. In this experiment, the face recognition model also manages to verify occluded faces despite slightly lower similarity scores. 
As presented in Fig.~\ref{fig:occlusion}, the CorrRISE algorithm precisely localizes the non-occluded regions that the model relies on for correct predictions and highlights the occlusions as dissimilar regions, such as masks and even another person's face.

From another perspective, the samples in Fig.~\ref{fig:verification} are selected from diverse demographic groups, e.g., various genders, ethnicities, and ages. The visual results validate that the CorrRISE method only depends on the decision of the recognition system and shows no significant bias across demographic groups.

\subsubsection{Challenging Verification Scenario}
Face verification systems can encounter various challenging situations in daily usage, such as head pose changes or very similar identities. 
It is important to provide reliable explanations for the system's behavior in specific scenarios. 

Head pose variation is a well-known challenge for face recognition systems. Fig.~\ref{fig:pose} shows four examples, where the model correctly recognizes the first two but fails at the last two. The saliency maps in Fig.~\ref{fig:pose1} indicate that the model manages to make correct predictions by localizing similar regions, such as cheeks and noses, with high saliency values even on the profile faces. 
In contrast, it fails to find enough similarities in the examples in Fig.~\ref{fig:pose2} and makes false predictions due to lacking sufficient information. For instance, the dissimilar saliency map spotlights the left eye of the front face in the third example, which corresponds to the missing parts in the profile face. 

Fig.~\ref{fig:fail} further presents two examples of similar identity scenarios in triplet format, where probe images are very similar to both the matching and non-matching gallery images. In both cases, the face recognition model has mistakenly verified the non-matching pair (right) as faces belonging to the same subject. 
The saliency maps generated by CorrRISE provide an explanation for this incorrect decision. These maps indicate that, despite having lower salient values compared to the matching pairs (left), the model perceives the nose and mouth regions of the two non-matching images are sufficiently close to classify them as the same person. Meanwhile, there is no significant dissimilar region between these images.

\subsubsection{Standard Identification Scenario}

Fig.~\ref{fig:idresults} presents two examples from the IJB-C dataset, each including a probe image and the top five most similar gallery images based on the prediction of the ArcFace face recognition model, see the first row. The similarity score between the probe image and each gallery image ranks from left to right. 

In the first example, the saliency map explanation produced by CorrRISE, see the last row, interprets that the model perceives a high similarity between the probe and gallery image. It also shows that the central face region of the second image, the nose of the third image, and the glasses of the fourth and fifth images are similar to the corresponding parts of the probe image, but to a lesser extent when compared to the first gallery image.
In the second example, CorrRISE reveals that the nose and beard areas of the first gallery image exhibit the highest similarity to the probe image. Overall, this visual explanation helps users understand why the face recognition model ranks the gallery images as it does, highlighting the specific regions that contribute to its decision.


\begin{figure*}[t]
	\centering
	\begin{adjustbox}{width=\linewidth}
    \includegraphics[]{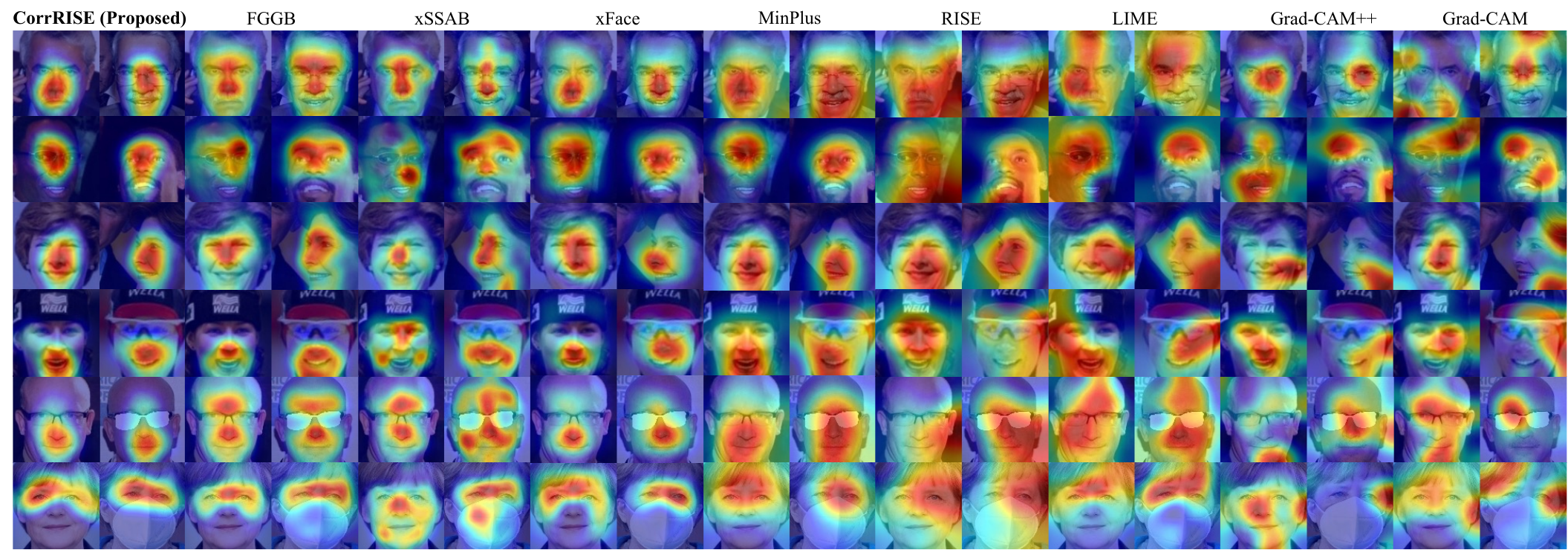}
	\end{adjustbox}
\caption{Visual results comparison among several saliency map-based explanation methods in face verification scenario. The importance increases from blue to red color.}
\label{fig:comparison}
\end{figure*}
\subsubsection{Comparison with Other Explanation Approaches}

A visual comparison is made between the proposed method and eight explanation approaches, including adaptations of four classical XAI techniques, i.e., Grad-CAM~\cite{selvaraju2017grad}, Grad-CAM++~\cite{chattopadhay2018grad}, LIME~\cite{ribeiro2016should}, and RISE~\cite{petsiuk2018rise}, and four state-of-the-art explanation methods in face recognition, namely MinPlus~\cite{mery2022true}, xFace~\cite{knoche2023explainable}, xSSAB~\cite{huber2024efficient}, and FGGB~\cite{lu2024explainable}. Implementation details of all the explanation methods refer to Section~\ref{label:implementation}.

Fig.~\ref{fig:comparison} and~\ref{fig:idresults} illustrate examples of saliency maps created by various explanation approaches under verification and identification scenarios respectively. In Fig.~\ref{fig:comparison}, only similar regions between two images are visualized and compared, due to the limitation of several earlier developed methods. The results show that CorrRISE consistently yields stable saliency maps, precisely highlighting the most similar regions between any given image pair. In comparison, the adapted Grad-CAM and Grad-CAM++ methods produce less stable and meaningful explanation maps. LIME and RISE tend to allocate a broad range of high-saliency pixels, making the importance map less precise. 
The other two perturbation-based methods, MinPlus and xFace, produce comparable visual results with CorrRISE, but the former fails to exclude mask regions (row 5 in Fig.~\ref{fig:comparison}). 
On the other hand, gradient backpropagation-based methods, FGGB and xSSAB, tend to provide more sparse salient regions and offer different explanations in certain examples (rows 4 and 5 in Fig.~\ref{fig:comparison}).

Fig.~\ref{fig:idresults} further compares the similarity maps under identification scenarios. It has been shown that different explanation methods yield various saliency maps as interpretations. For instance in the second example of Fig.~\ref{fig:idresults}, CorrRISE and xFace highlight the nose and beard areas of the rank-1 gallery image as high similarity regions, while gradient-based methods pay more attention to eyebrows (xSSAB) or nose (FGGB).
Nevertheless, determining which method provides more accurate saliency maps than others is challenging based solely on visualization results. 
Thus, quantitative evaluation with objective metrics is performed and the results are reported in the follow-up section.

\begin{table}[t]
  \centering
  \caption{Quantitative evaluation of similarity maps in face verification scenario. Deletion and Insertion metrics (\%) are reported on LFW, CPLFW, and CALFW datasets, representing three different verification scenarios. \textbf{Del~($\downarrow$)} refers to the Deletion metric, the smaller the better. \textbf{Ins~($\uparrow$)} refers to the Insertion metric, the larger the better. }
    \resizebox{0.575\textwidth}{!}{
    \begin{tabular}{c|cc|cc|cc}
    \toprule
    \multirow{2}[4]{*}{Methods} & \multicolumn{2}{c|}{LFW} & \multicolumn{2}{c|}{CPLFW} & \multicolumn{2}{c}{CALFW} \\
\cmidrule{2-7}          & Del   & Ins   & Del   & Ins   & Del   & Ins \\
    \midrule
    Grad-CAM\cite{selvaraju2017grad} & 50.00 & 68.90 & 39.53 & 59.94 & 47.19 & 69.32 \\
    Grad-CAM++\cite{chattopadhay2018grad} & 45.05 & 72.64 & 36.16 & 62.06 & 44.23 & 71.20 \\
    LIME\cite{ribeiro2016should}  & 35.71 & 80.76 & 26.61 & 72.69 & 34.03 & 78.59 \\
    RISE\cite{petsiuk2018rise}  & 34.46 & 82.82 & 31.32 & 68.17 & 32.40  & 80.43   \\
    \midrule
    MinPlus\cite{mery2022true} & 29.74 & 83.49 & 24.58 & 69.76 & 29.43 & 79.85 \\
    xFace\cite{knoche2023explainable} & 25.54 & 87.13 & 20.66 & 76.68 & 24.21 & 83.78 \\
    xSSAB\cite{huber2024efficient} & 27.02 & 83.69 & 23.25 & 71.88 & 27.12 & 79.50\\
    FGGB\cite{lu2024explainable} & 25.67 & 84.43 & 21.41 & 73.04 & 26.28 & 79.98\\
    \midrule
    CorrRISE (Proposed) & \textbf{23.37} & \textbf{87.15} & \textbf{18.01} & \textbf{78.50} & \textbf{22.62} & \textbf{83.82} \\
    \bottomrule
    \end{tabular}%
    }
  \label{tab:simEval}%
\end{table}%

\begin{table}[t]
  \centering
  \caption{Quantitative evaluation of dissimilarity maps in face verification scenario. Deletion and Insertion metrics (\%) are reported on LFW, CPLFW, and Webface-OCC datasets, representing three different verification scenarios. \textbf{Del~($\downarrow$)} refers to the Deletion metric, the smaller the better. \textbf{Ins~($\uparrow$)} refers to the Insertion metric, the larger the better. }
    \resizebox{0.575\textwidth}{!}{
    \begin{tabular}{c|cc|cc|cc}
    \toprule
    \multirow{2}[4]{*}{Methods} & \multicolumn{2}{c|}{LFW} & \multicolumn{2}{c|}{CPLFW} & \multicolumn{2}{c}{CALFW} \\
\cmidrule{2-7}          & Del   & Ins   & Del   & Ins   & Del   & Ins \\
    \midrule
    xFace\cite{knoche2023explainable} & 75.59 & 92.70 & 50.83 & 87.65 & 62.61 & \textbf{91.20} \\
    xSSAB\cite{huber2024efficient}  & 48.06 & \textbf{93.41} & 32.51 & 88.10 & 40.01 & 89.75 \\
    FGGB\cite{lu2024explainable}  & \textbf{44.99} & 93.06 & \textbf{29.01} & \textbf{88.26} & \textbf{36.87} & 88.86 \\
    \midrule
    CorrRISE (Proposed) & 79.64 & 90.23 & 53.70 & 84.21 & 64.77 & 87.60 \\
    \bottomrule
    \end{tabular}%
    }
  \label{tab:dissimEval}%
\end{table}%

\subsection{Quantitative Evaluation Results}
\label{Section4-4}
This section reports the ``Deletion'' and ``Insertion'' metrics for the quantitative evaluation of various explanation approaches. As described in Section~\ref{chapter:eval}, these metrics measure the changes in recognition performance after modifying the input images according to the saliency map. 

Table~\ref{tab:simEval} first demonstrates the evaluation results of similarity maps generated by the state-of-the-art explanation methods in three typical face verification scenarios. Generally, the reported metrics are consistent with the visual observations in Fig.~\ref{fig:comparison}. For example, the adapted Grad-CAM and Grad-CAM++ show poorer performance than other state-of-the-art explanation methods. Notably, CorrRISE achieves much better quantitative results than a straightforward adaptation of RISE. The latter can only obtain comparable results with other adapted XAI methods such as LIME, demonstrating the advantage and value of our proposed CorrRISE method. Although the recent explainable face verification methods present some visually compelling results, quantitative metrics show that CorrRISE is capable of providing more precise saliency maps on all three testing datasets. It is also interesting to observe that perturbation-based methods, such as xFace and CorrRISE, generally obtain more precise similarity maps than gradient-based methods, i.e., xSSAB and FGGB. 
On the other hand, Table~\ref{tab:dissimEval} reports the evaluation results on dissimilarity maps in non-matching cases. It involves only four explanation methods as other approaches cannot deal with non-matching cases. The results show that the gradient backpropagation mechanism leads to better performance in generating dissimilarity maps.

\begin{table}[t]
  \centering
  \caption{Quantitative evaluation of similarity maps in face identification scenario. Deletion and Insertion metrics (\%) are reported on IJB-C datasets. \textbf{Del~($\downarrow$)} refers to the Deletion metric, the smaller the better. \textbf{Ins~($\uparrow$)} refers to the Insertion metric, the larger the better. }
    \resizebox{0.48\textwidth}{!}{
    \begin{tabular}{c|cc|cc}
    \toprule
    \multirow{2}[4]{*}{Methods} & \multicolumn{2}{c|}{IJB-C (Rank-1)} & \multicolumn{2}{c}{IJB-C (Rank-5)}  \\
\cmidrule{2-5}          & Del   & Ins   & Del   & Ins  \\
    \midrule
    LIME\cite{ribeiro2016should} & 21.29 & 59.50 & 24.85 & 65.05 \\
    MinPlus\cite{mery2022true} & 17.38 & 54.80 & 21.33 & 60.47 \\ 
    xFace\cite{knoche2023explainable} & 15.16 & 64.23 & 18.52 & 69.79  \\
    xSSAB\cite{huber2024efficient} & 14.95 & 63.48 & 18.17 & 68.61 \\
    FGGB\cite{lu2024explainable} & 14.67 & 63.58 & 17.87 & 68.71 \\
    \midrule
    CorrRISE (Proposed) & \textbf{14.30} & \textbf{64.81} & \textbf{17.43} & \textbf{72.10} \\
    \bottomrule
    \end{tabular}%
    }
  \label{tab:idenEval}%
\end{table}%

\begin{table*}[t]
  \centering
  \caption{Extra experiments on the types of masks and the number of iterations. The notations are explained as follows. BM: Binary Mask, RdM: Random Mask, GaussM: Gaussian Mask, N1000: 1000 iterations. The default configuration is BM+N1000. \textbf{Del~($\downarrow$)} refers to the Deletion metric, the smaller the better. \textbf{Ins~($\uparrow$)} refers to the Insertion metric, the larger the better. \textcolor[rgb]{ 1,  0,  0}{Red color} denotes the highest score and \textcolor[rgb]{ .267,  .447,  .769}{blue color} denotes the second highest score. }
  \resizebox{\linewidth}{!}{%
    \begin{tabular}{c|cccc|cccc|cccc}
    \toprule
    \multirow{3}[6]{*}{Configurations} & \multicolumn{4}{c|}{LFW}      & \multicolumn{4}{c|}{CPLFW}    & \multicolumn{4}{c}{CALFW} \\
\cmidrule{2-13}          & \multicolumn{2}{c}{Similarity} & \multicolumn{2}{c|}{Dissimilarity} & \multicolumn{2}{c}{Similarity} & \multicolumn{2}{c|}{Dissimilarity} & \multicolumn{2}{c}{Similarity} & \multicolumn{2}{c}{Dissimilarity} \\
\cmidrule{2-13}          & Del   & Ins   & Del   & Ins   & Del   & Ins   & Del   & Ins   & Del   & Ins   & Del   & Ins \\
    \midrule
    BM+N1000 & \textcolor[rgb]{ .267,  .447,  .769}{23.37} & \textcolor[rgb]{ 1,  0,  0}{87.15} & \textcolor[rgb]{ .267,  .447,  .769}{79.64} & \textcolor[rgb]{ .267,  .447,  .769}{90.23} & \textcolor[rgb]{ .267,  .447,  .769}{18.01} & \textcolor[rgb]{ 1,  0,  0}{78.50} & \textcolor[rgb]{ .267,  .447,  .769}{53.70} & \textcolor[rgb]{ .267,  .447,  .769}{84.21} & \textcolor[rgb]{ .267,  .447,  .769}{22.62} & 83.82 & \textcolor[rgb]{ .267,  .447,  .769}{64.77} & 87.60 \\
    RdM+N1000 & 23.96 & \textcolor[rgb]{ .267,  .447,  .769}{87.04} & 81.22 & 90.13 & 18.75 & 78.21 & 55.89 & 83.57 & 23.49 & 83.64 & 66.66 & 87.40 \\
    GaussM+N1000 & 24.61 & 86.60 & 80.50  & \textcolor[rgb]{ 1,  0,  0}{90.94} & 19.40 & 77.89 & 54.42 & \textcolor[rgb]{ 1,  0,  0}{84.71} & 23.70 & 83.53 & 65.76 & \textcolor[rgb]{ 1,  0,  0}{88.75} \\
    BM+N100 & 24.76 & 86.95 & 80.98 & 88.63 & 18.97 & 77.69 & 56.78 & 82.08 & 24.03 & 83.58 & 68.21 & 86.12 \\
    BM+N500 & 23.58 & \textcolor[rgb]{ 1,  0,  0}{87.15} & 80.17 & 89.93 & 18.07 & 78.43 & 54.29 & 83.94 & 22.74 & \textcolor[rgb]{ 1,  0,  0}{83.88} & 65.78 & 87.27 \\
    BM+N1500 & \textcolor[rgb]{ 1,  0,  0}{23.35} & \textcolor[rgb]{ 1,  0,  0}{87.15} & \textcolor[rgb]{ 1,  0,  0}{79.45} & 90.16 & \textcolor[rgb]{ 1,  0,  0}{17.95} & \textcolor[rgb]{ .267,  .447,  .769}{78.48} & \textcolor[rgb]{ 1,  0,  0}{53.47} & 84.07 & \textcolor[rgb]{ 1,  0,  0}{22.55} & \textcolor[rgb]{ .267,  .447,  .769}{83.86} & \textcolor[rgb]{ 1,  0,  0}{64.24} & \textcolor[rgb]{ .267,  .447,  .769}{87.67} \\
    \bottomrule
    \end{tabular}%
    }
  \label{tab:corrrise-ablation}%
\end{table*}%

In the face identification scenario, we leverage rank-1 and rank-5 identification on the IJB-C dataset as an auxiliary task and perform similar ``Deletion'' and ``Insertion'' processes to evaluate the similarity maps created by different explanation methods. As shown in Table~\ref{tab:idenEval}, CorrRISE consistently achieves superior results in both metrics, which implies that, in Fig.~\ref{fig:idresults}, the saliency maps provided by CorrRISE better highlight the critical pixels than others for the face recognition system.

Furthermore, additional experiments are conducted on the hyperparameters of the CorrRISE algorithm, namely mask types and number of iterations. The default configuration employs binary masks as perturbations and runs 1000 iterations for the saliency map generation process. Table~\ref{tab:corrrise-ablation} shows that applying binary masks typically outperforms the other types of occlusions, such as masks with randomized or Gaussian distributed values, although Gaussian masks provide slightly better results in the insertion scores for dissimilarity maps. Apart from mask types, the number of iterations also presents a substantial impact on the performance of the CorrRISE algorithm. In general, a larger number of iterations leads to better results. For example, the saliency maps after running 1000 iterations are notably more accurate than 100 iterations. However, there is a trade-off between performance and efficiency, as running more iterations also requires more processing time. 

\begin{table}[t]
  \centering
  \caption{Explainablity performance of CorrRISE tested on different state-of-the-art face recognition models. The verification accuracy (\%) of face recognition models and two explainability metrics (\%) for CorrRISE are reported.}
    \resizebox{0.55\linewidth}{!}{%
    \begin{tabular}{c|c|cc}
    \toprule
    Models & Acc (LFW) & Deletion ($\downarrow$) & Insertion ($\uparrow$) \\
    \midrule
    ArcFace~\cite{deng2019arcface} & 99.53 & 23.37 & 87.15 \\
    MagFace~\cite{meng2021magface} & 99.83 & 23.35 & 85.68  \\
    AdaFace~\cite{kim2022adaface} & 99.82 & 23.31 & 85.63 \\
    \bottomrule
    \end{tabular}%
    }
  \label{tab:crossmodel}%
\end{table}%

\begin{figure*}[t]
	\centering
	\begin{adjustbox}{width=\linewidth}
    \includegraphics[]{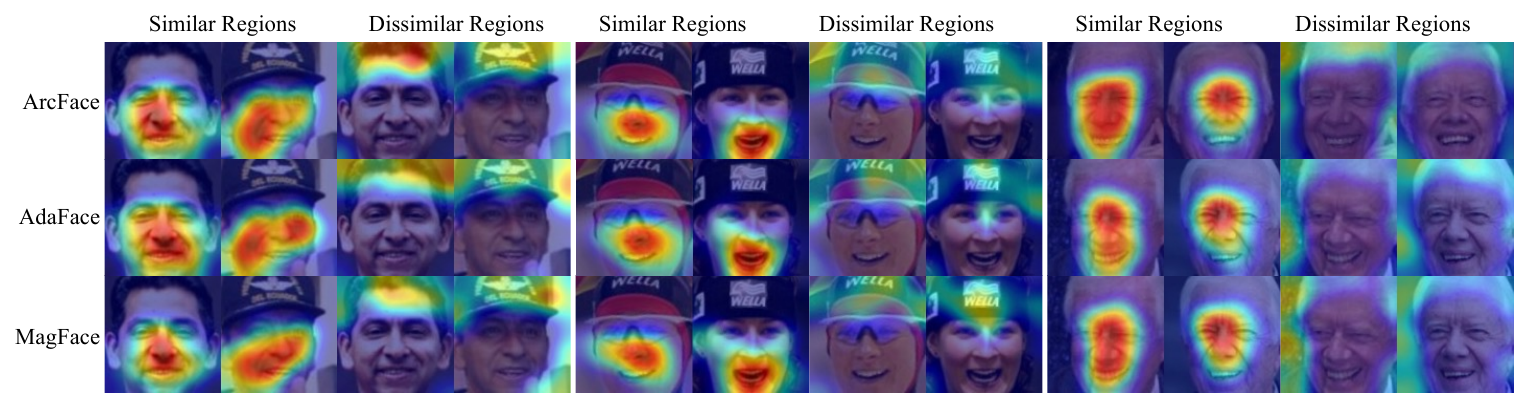}
	\end{adjustbox}
\caption{Visual saliency maps produced by CorrRISE for three different deep face recognition models. The visualization results are consistent with the quantitative metrics reported in the manuscript.}
\label{fig:FRmodel}
\end{figure*}
\begin{figure*}[t]
	\centering
	\begin{adjustbox}{width=0.8\linewidth}
    \includegraphics[]{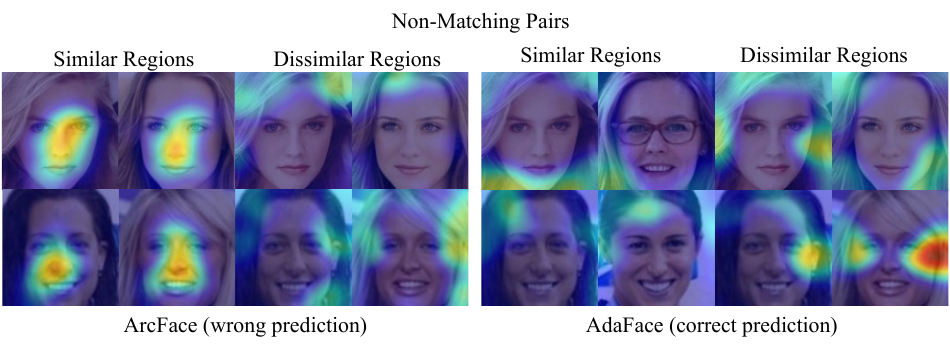}
	\end{adjustbox}
\caption{Visual explanation of two different decisions made by two face recognition models. ArcFace model mistakenly verifies the given example as matching while the AdaFace correctly recognizes they are non-matching. }
\label{fig:FRmodel2}
\end{figure*}

\subsection{Generalization across Different Face Recognition Models}
CorrRISE is additionally applied to different face recognition models to showcase the generalization ability of our proposed method. The verification accuracy of these models and the explainability performance of CorrRISE are reported in Table~\ref{tab:crossmodel}. In principle, the explainability metrics are used to compare explanation methods tested on the same face recognition model and the scores are not directly comparable across different models. However, the table also shows that when the models achieve similar verification accuracy on the LFW dataset, the generated saliency maps also report similar scores, which validates that CorrRISE is model-agnostic.

Fig.~\ref{fig:FRmodel} further provides visualization samples of the saliency maps that CorrRISE generated for three face recognition models. The saliency maps highlight similar regions across models, which validates that our proposed metrics are consistent with the visualization results and are reliable for a fair comparison among general saliency map-based explanation methods. It also shows that CorrRISE generalizes well across different face recognition models.

In Fig.~\ref{fig:FRmodel}, the produced saliency maps are very similar because all face recognition models make correct predictions on the three examples. 
To further show the strong explainability of CorrRISE, Fig.~\ref{fig:FRmodel2} provides visual explanations for two opposite decisions made by ArcFace~\cite{deng2019arcface} and AdaFace~\cite{kim2022adaface} respectively. The former fails to recognize the given example and mistakenly verifies they are the same person, while the latter makes correct predictions. The saliency map produced by the CorrRISE method shows that the ArcFace model allocates high saliency values to similar regions between the non-matching pairs while indicating very low dissimilarity between them. 
On the contrary, the AdaFace model makes correct predictions because it believes there are strong dissimilar pixels between the given examples. 
This experiment is complementary to Fig.~\ref{fig:FRmodel}, and further proves that CorrRISE is capable of providing meaningful explanations for different face recognition models.


\begin{table}[t]
  \centering
  \caption{Quantitative evaluation of similarity maps generated by CorrRISE with regularization. Deletion and Insertion metrics (\%) are reported on LFW, CPLFW, and CALFW datasets. \textbf{Del~($\downarrow$)} refers to the Deletion metric, the smaller the better. \textbf{Ins~($\uparrow$)} refers to the Insertion metric, the larger the better. \textcolor[rgb]{ 1,  0,  0}{Red color} denotes the highest score and \textcolor[rgb]{ .267,  .447,  .769}{blue color} denotes the second highest score. }
    \resizebox{0.6\textwidth}{!}{
    \begin{tabular}{c|cc|cc|cc}
    \toprule
    \multirow{2}[4]{*}{Methods} & \multicolumn{2}{c|}{LFW} & \multicolumn{2}{c|}{CPLFW} & \multicolumn{2}{c}{CALFW} \\
\cmidrule{2-7}          & Del   & Ins   & Del   & Ins   & Del   & Ins \\
    \midrule
    xFace\cite{knoche2023explainable} & 25.54 & 87.13 & 20.66 & 76.68 & 24.21 & \textcolor[rgb]{ .267,  .447,  .769}{83.78} \\
    xSSAB\cite{huber2024efficient} & 27.02 & 83.69 & 23.25 & 71.88 & 27.12 & 79.50\\
    FGGB\cite{lu2024explainable} & 25.67 & 84.43 & 21.41 & 73.04 & 26.28 & 79.98\\
    \midrule
    CorrRISE (Proposed) & \textcolor[rgb]{ 1,  0,  0}{23.37} & 87.15 & \textcolor[rgb]{ 1,  0,  0}{18.01} & \textcolor[rgb]{ 1,  0,  0}{78.50} & \textcolor[rgb]{ 1,  0,  0}{22.62} & \textcolor[rgb]{ 1,  0,  0}{83.82} \\
    +Regularization & \textcolor[rgb]{ .267,  .447,  .769}{23.46} & \textcolor[rgb]{ 1,  0,  0}{87.17} & \textcolor[rgb]{ .267,  .447,  .769}{18.62} & \textcolor[rgb]{ .267,  .447,  .769}{77.28} & \textcolor[rgb]{ .267,  .447,  .769}{23.09} & 83.56 \\
    \bottomrule
    \end{tabular}%
    }
  \label{tab:simEval-Regularize}%
\end{table}%

\begin{table}[t]
  \centering
  \caption{Quantitative evaluation of dissimilarity maps generated by CorrRISE with regularization.  Deletion and Insertion metrics (\%) are reported on LFW, CPLFW, and CALFW datasets. \textbf{Del~($\downarrow$)} refers to the Deletion metric, the smaller the better. \textbf{Ins~($\uparrow$)} refers to the Insertion metric, the larger the better. \textcolor[rgb]{ 1,  0,  0}{Red color} denotes the highest score and \textcolor[rgb]{ .267,  .447,  .769}{blue color} denotes the second highest score. }
    \resizebox{0.6\textwidth}{!}{
    \begin{tabular}{c|cc|cc|cc}
    \toprule
    \multirow{2}[4]{*}{Methods} & \multicolumn{2}{c|}{LFW} & \multicolumn{2}{c|}{CPLFW} & \multicolumn{2}{c}{CALFW} \\
\cmidrule{2-7}          & Del   & Ins   & Del   & Ins   & Del   & Ins \\
    \midrule
    xFace\cite{knoche2023explainable} & 75.59 & 92.70 & 50.83 & 87.65 & 62.61 & \textcolor[rgb]{ .267,  .447,  .769}{91.20} \\
    xSSAB\cite{huber2024efficient}  & \textcolor[rgb]{ .267,  .447,  .769}{48.06} & \textcolor[rgb]{ .267,  .447,  .769}{93.41} & \textcolor[rgb]{ .267,  .447,  .769}{32.51} & \textcolor[rgb]{ .267,  .447,  .769}{88.10} & \textcolor[rgb]{ .267,  .447,  .769}{40.01} & 89.75 \\
    FGGB\cite{lu2024explainable}  & \textcolor[rgb]{ 1,  0,  0}{44.99} & 93.06 & \textcolor[rgb]{ 1,  0,  0}{29.01} & \textcolor[rgb]{ 1,  0,  0}{88.26} & \textcolor[rgb]{ 1,  0,  0}{36.87} & 88.86 \\
    \midrule
    CorrRISE (Proposed) & 79.64 & 90.23 & 53.70 & 84.21 & 64.77 & 87.60 \\
    +Regularization & 63.65 & \textcolor[rgb]{ 1,  0,  0}{93.60} & 50.18 & 87.70 & 51.28 & \textcolor[rgb]{ 1,  0,  0}{92.12} \\
    \bottomrule
    \end{tabular}%
    }
  \label{tab:dissimEval-Regularize}%
\end{table}%

\subsection{Limitation and Improvement}
As shown in Section~\ref{Section4-4}, although CorrRISE achieves superior performance in generating similarity maps, it shows relatively poor results on dissimilarity maps when compared to gradient-based methods xSSAB and FGGB. The other perturbation-based method, xFace, falls into a similar situation. This is a potential limitation of perturbation-based explanation methods when they are applied to vision tasks involving comparisons between images, such as face recognition, image retrieval, etc.
In principle, perturbation-based methods first inject occlusions into an input image and observe the impact on output. In the case of face recognition, the output is a cosine similarity score, and the impact is typically measured as the change in the similarity score. For matching image pairs, the original similarity score is already high and the subsequent change in it after injecting perturbations is obvious to detect. Nevertheless, for non-matching pairs with naturally low similarity scores, the perturbations added to them bring very limited impact to the output, thus resulting in less accurate dissimilarity maps.

This section proposes a solution to alleviate this problem by introducing an additional regularization term to a specific step in the proposed CorrRISE algorithm. The idea is to reformulate the cosine similarity score when calculating dissimilarity maps for non-matching images. Specifically, we first replace the unmasked area of $I_A$ with corresponding pixels from $I_B$ to obtain $I_A^{reg}$, namely:
\begin{align}
  I_A^{reg} = I_A \odot (1-M_i) + I_B \odot M_i.
\end{align}
It is fed to the feature extractor to get deep feature representation $x_A^{reg}$. Then, the cosine similarity between $x_A^{reg}$ and $x_B$ is calculated. Ideally, when the random mask occludes the most dissimilar regions of $\{I_A, I_B\}$, the similarity score of $\{x_A^{reg}, x_B\}$ will change more significantly than that of $\{x_A^m, x_B\}$.
Finally, the computation of $SC_A$ in Algorithm~\ref{alg:corrrise} is reformulated as follows:
\begin{align}
    \lambda = \frac{\texttt{Score}(x_A, x_B)-1}{\texttt{Score}(x_A, x_B)+1},
\end{align}
\begin{align}
    SC_A[k] = \texttt{Score}(x_A^m, x_B) + \lambda \cdot  \texttt{Score}(x_A^{reg}, x_B),
\end{align}
where $\lambda$ varies between $[-1,0]$ as the similarity score of $\{x_A, x_B\}$ varies between $[0,1]$. The lower cosine similarity score between $\{I_A, I_B\}$, the more weights allocating to the regularization term.

As shown in Table~\ref{tab:simEval-Regularize} and~\ref{tab:dissimEval-Regularize}, CorrRISE with regularization achieves significant improvement compared to other perturbation-based methods in generating dissimilarity maps, with some of the insertion scores even beating the state of the arts. Moreover, despite a subtle decline, its performance in similarity maps remains superior to other explanation methods.
Overall, this section provides a direction to address the limitation of general perturbation-based methods in identifying dissimilar regions between images. 

\section{Conclusion}

This manuscript presented a significant step forward in the field of explainable face recognition by contributing a comprehensive explanation framework. It broadened the application of visual saliency map-based explanations to the most common face verification and identification tasks and provided fair definitions for this problem. The framework provided a model-agnostic explanation algorithm, called CorrRISE. It addressed the explainability problem in face recognition systems by generating saliency maps that highlight both similar and dissimilar regions between given face images. Extensive visualization results in multiple scenarios demonstrated the advantage of our proposed method and showcased its capability to analyze potential failure cases in challenging recognition scenarios and, thereby provided insights into improving the current face recognition system. Furthermore, a new evaluation methodology was designed, constituting a key component in the explanation framework. This methodology offered a quantitative assessment and fair comparison for general saliency map-based explainable face recognition approaches and would benefit future research in this area. The quantitative assessment results using this new evaluation methodology demonstrated the state-of-the-art performance of CorrRISE in generating similarity maps between face images. 
Lastly, this manuscript exploited a limitation of general perturbation-based methods in identifying dissimilar regions between images and provided a tentative solution to improve them.

\bibliographystyle{ACM-Reference-Format}
\bibliography{egbib}


\begin{thebibliography}{47}


\ifx \showCODEN    \undefined \def \showCODEN     #1{\unskip}     \fi
\ifx \showDOI      \undefined \def \showDOI       #1{#1}\fi
\ifx \showISBNx    \undefined \def \showISBNx     #1{\unskip}     \fi
\ifx \showISBNxiii \undefined \def \showISBNxiii  #1{\unskip}     \fi
\ifx \showISSN     \undefined \def \showISSN      #1{\unskip}     \fi
\ifx \showLCCN     \undefined \def \showLCCN      #1{\unskip}     \fi
\ifx \shownote     \undefined \def \shownote      #1{#1}          \fi
\ifx \showarticletitle \undefined \def \showarticletitle #1{#1}   \fi
\ifx \showURL      \undefined \def \showURL       {\relax}        \fi
\providecommand\bibfield[2]{#2}
\providecommand\bibinfo[2]{#2}
\providecommand\natexlab[1]{#1}
\providecommand\showeprint[2][]{arXiv:#2}

\bibitem[Adebayo et~al\mbox{.}(2018)]%
        {adebayo2018sanity}
\bibfield{author}{\bibinfo{person}{Julius Adebayo}, \bibinfo{person}{Justin Gilmer}, \bibinfo{person}{Michael Muelly}, \bibinfo{person}{Ian Goodfellow}, \bibinfo{person}{Moritz Hardt}, {and} \bibinfo{person}{Been Kim}.} \bibinfo{year}{2018}\natexlab{}.
\newblock \showarticletitle{Sanity checks for saliency maps}.
\newblock \bibinfo{journal}{\emph{Advances in neural information processing systems}}  \bibinfo{volume}{31} (\bibinfo{year}{2018}).
\newblock


\bibitem[Bau et~al\mbox{.}(2017)]%
        {bau2017network}
\bibfield{author}{\bibinfo{person}{David Bau}, \bibinfo{person}{Bolei Zhou}, \bibinfo{person}{Aditya Khosla}, \bibinfo{person}{Aude Oliva}, {and} \bibinfo{person}{Antonio Torralba}.} \bibinfo{year}{2017}\natexlab{}.
\newblock \showarticletitle{Network dissection: Quantifying interpretability of deep visual representations}. In \bibinfo{booktitle}{\emph{Proceedings of the IEEE conference on computer vision and pattern recognition}}. \bibinfo{pages}{6541--6549}.
\newblock


\bibitem[Binder et~al\mbox{.}(2016)]%
        {binder2016layer}
\bibfield{author}{\bibinfo{person}{Alexander Binder}, \bibinfo{person}{Gr{\'e}goire Montavon}, \bibinfo{person}{Sebastian Lapuschkin}, \bibinfo{person}{Klaus-Robert M{\"u}ller}, {and} \bibinfo{person}{Wojciech Samek}.} \bibinfo{year}{2016}\natexlab{}.
\newblock \showarticletitle{Layer-wise relevance propagation for neural networks with local renormalization layers}. In \bibinfo{booktitle}{\emph{Artificial Neural Networks and Machine Learning--ICANN 2016: 25th International Conference on Artificial Neural Networks, Barcelona, Spain, September 6-9, 2016, Proceedings, Part II 25}}. Springer, \bibinfo{pages}{63--71}.
\newblock


\bibitem[Castanon and Byrne(2018)]%
        {castanon2018visualizing}
\bibfield{author}{\bibinfo{person}{Gregory Castanon} {and} \bibinfo{person}{Jeffrey Byrne}.} \bibinfo{year}{2018}\natexlab{}.
\newblock \showarticletitle{Visualizing and quantifying discriminative features for face recognition}. In \bibinfo{booktitle}{\emph{2018 13th IEEE International Conference on Automatic Face \& Gesture Recognition (FG 2018)}}. IEEE, \bibinfo{pages}{16--23}.
\newblock


\bibitem[Chattopadhay et~al\mbox{.}(2018)]%
        {chattopadhay2018grad}
\bibfield{author}{\bibinfo{person}{Aditya Chattopadhay}, \bibinfo{person}{Anirban Sarkar}, \bibinfo{person}{Prantik Howlader}, {and} \bibinfo{person}{Vineeth~N Balasubramanian}.} \bibinfo{year}{2018}\natexlab{}.
\newblock \showarticletitle{Grad-cam++: Generalized gradient-based visual explanations for deep convolutional networks}. In \bibinfo{booktitle}{\emph{2018 IEEE winter conference on applications of computer vision (WACV)}}. IEEE, \bibinfo{pages}{839--847}.
\newblock


\bibitem[Dabkowski and Gal(2017)]%
        {dabkowski2017real}
\bibfield{author}{\bibinfo{person}{Piotr Dabkowski} {and} \bibinfo{person}{Yarin Gal}.} \bibinfo{year}{2017}\natexlab{}.
\newblock \showarticletitle{Real time image saliency for black box classifiers}.
\newblock \bibinfo{journal}{\emph{Advances in neural information processing systems}}  \bibinfo{volume}{30} (\bibinfo{year}{2017}).
\newblock


\bibitem[Deng et~al\mbox{.}(2019)]%
        {deng2019arcface}
\bibfield{author}{\bibinfo{person}{Jiankang Deng}, \bibinfo{person}{Jia Guo}, \bibinfo{person}{Niannan Xue}, {and} \bibinfo{person}{Stefanos Zafeiriou}.} \bibinfo{year}{2019}\natexlab{}.
\newblock \showarticletitle{Arcface: Additive angular margin loss for deep face recognition}. In \bibinfo{booktitle}{\emph{Proceedings of the IEEE/CVF conference on computer vision and pattern recognition}}. \bibinfo{pages}{4690--4699}.
\newblock


\bibitem[Dong et~al\mbox{.}(2019)]%
        {dong2019explainability}
\bibfield{author}{\bibinfo{person}{Bo Dong}, \bibinfo{person}{Roddy Collins}, {and} \bibinfo{person}{Anthony Hoogs}.} \bibinfo{year}{2019}\natexlab{}.
\newblock \showarticletitle{Explainability for Content-Based Image Retrieval.}. In \bibinfo{booktitle}{\emph{CVPR Workshops}}. \bibinfo{pages}{95--98}.
\newblock


\bibitem[Elharrouss et~al\mbox{.}(2020)]%
        {elharrouss2020lfr}
\bibfield{author}{\bibinfo{person}{Omar Elharrouss}, \bibinfo{person}{Noor Almaadeed}, {and} \bibinfo{person}{Somaya Al-Maadeed}.} \bibinfo{year}{2020}\natexlab{}.
\newblock \showarticletitle{LFR face dataset: Left-Front-Right dataset for pose-invariant face recognition in the wild}. In \bibinfo{booktitle}{\emph{2020 IEEE International Conference on Informatics, IoT, and Enabling Technologies (ICIoT)}}. IEEE, \bibinfo{pages}{124--130}.
\newblock


\bibitem[Fong and Vedaldi(2017)]%
        {fong2017interpretable}
\bibfield{author}{\bibinfo{person}{Ruth~C Fong} {and} \bibinfo{person}{Andrea Vedaldi}.} \bibinfo{year}{2017}\natexlab{}.
\newblock \showarticletitle{Interpretable explanations of black boxes by meaningful perturbation}. In \bibinfo{booktitle}{\emph{Proceedings of the IEEE international conference on computer vision}}. \bibinfo{pages}{3429--3437}.
\newblock


\bibitem[Guo et~al\mbox{.}(2016)]%
        {guo2016ms-celeb-1m}
\bibfield{author}{\bibinfo{person}{Yandong Guo}, \bibinfo{person}{Lei Zhang}, \bibinfo{person}{Yuxiao Hu}, \bibinfo{person}{Xiaodong He}, {and} \bibinfo{person}{Jianfeng Gao}.} \bibinfo{year}{2016}\natexlab{}.
\newblock \showarticletitle{Ms-celeb-1m: {A} dataset and benchmark for large-scale face recognition}. In \bibinfo{booktitle}{\emph{Computer {Vision}–{ECCV} 2016: 14th {European} {Conference}, {Amsterdam}, {The} {Netherlands}, {October} 11-14, 2016, {Proceedings}, {Part} {III} 14}}. \bibinfo{pages}{87--102}.
\newblock


\bibitem[He et~al\mbox{.}(2016)]%
        {he2016deep}
\bibfield{author}{\bibinfo{person}{Kaiming He}, \bibinfo{person}{Xiangyu Zhang}, \bibinfo{person}{Shaoqing Ren}, {and} \bibinfo{person}{Jian Sun}.} \bibinfo{year}{2016}\natexlab{}.
\newblock \showarticletitle{Deep residual learning for image recognition}. In \bibinfo{booktitle}{\emph{Proceedings of the IEEE conference on computer vision and pattern recognition}}. \bibinfo{pages}{770--778}.
\newblock


\bibitem[Herman(2017)]%
        {herman2017promise}
\bibfield{author}{\bibinfo{person}{Bernease Herman}.} \bibinfo{year}{2017}\natexlab{}.
\newblock \showarticletitle{The promise and peril of human evaluation for model interpretability}.
\newblock \bibinfo{journal}{\emph{arXiv preprint arXiv:1711.07414}} (\bibinfo{year}{2017}).
\newblock


\bibitem[Hu et~al\mbox{.}(2022)]%
        {hu2022x}
\bibfield{author}{\bibinfo{person}{Brian Hu}, \bibinfo{person}{Bhavan Vasu}, {and} \bibinfo{person}{Anthony Hoogs}.} \bibinfo{year}{2022}\natexlab{}.
\newblock \showarticletitle{X-mir: Explainable medical image retrieval}. In \bibinfo{booktitle}{\emph{Proceedings of the IEEE/CVF Winter Conference on Applications of Computer Vision}}. \bibinfo{pages}{440--450}.
\newblock


\bibitem[Huang et~al\mbox{.}(2021)]%
        {huang2021face}
\bibfield{author}{\bibinfo{person}{Baojin Huang}, \bibinfo{person}{Zhongyuan Wang}, \bibinfo{person}{Guangcheng Wang}, \bibinfo{person}{Kui Jiang}, \bibinfo{person}{Kangli Zeng}, \bibinfo{person}{Zhen Han}, \bibinfo{person}{Xin Tian}, {and} \bibinfo{person}{Yuhong Yang}.} \bibinfo{year}{2021}\natexlab{}.
\newblock \showarticletitle{When face recognition meets occlusion: A new benchmark}. In \bibinfo{booktitle}{\emph{ICASSP 2021-2021 IEEE International Conference on Acoustics, Speech and Signal Processing (ICASSP)}}. IEEE, \bibinfo{pages}{4240--4244}.
\newblock


\bibitem[Huang et~al\mbox{.}(2008)]%
        {huang2008labeled}
\bibfield{author}{\bibinfo{person}{Gary~B Huang}, \bibinfo{person}{Marwan Mattar}, \bibinfo{person}{Tamara Berg}, {and} \bibinfo{person}{Eric Learned-Miller}.} \bibinfo{year}{2008}\natexlab{}.
\newblock \showarticletitle{Labeled faces in the wild: A database forstudying face recognition in unconstrained environments}. In \bibinfo{booktitle}{\emph{Workshop on faces in'Real-Life'Images: detection, alignment, and recognition}}.
\newblock


\bibitem[Huber et~al\mbox{.}(2024)]%
        {huber2024efficient}
\bibfield{author}{\bibinfo{person}{Marco Huber}, \bibinfo{person}{Anh~Thi Luu}, \bibinfo{person}{Philipp Terh{\"o}rst}, {and} \bibinfo{person}{Naser Damer}.} \bibinfo{year}{2024}\natexlab{}.
\newblock \showarticletitle{Efficient explainable face verification based on similarity score argument backpropagation}. In \bibinfo{booktitle}{\emph{Proceedings of the IEEE/CVF Winter Conference on Applications of Computer Vision}}. \bibinfo{pages}{4736--4745}.
\newblock


\bibitem[Kim et~al\mbox{.}(2022)]%
        {kim2022adaface}
\bibfield{author}{\bibinfo{person}{Minchul Kim}, \bibinfo{person}{Anil~K Jain}, {and} \bibinfo{person}{Xiaoming Liu}.} \bibinfo{year}{2022}\natexlab{}.
\newblock \showarticletitle{Adaface: Quality adaptive margin for face recognition}. In \bibinfo{booktitle}{\emph{Proceedings of the IEEE/CVF Conference on Computer Vision and Pattern Recognition}}. \bibinfo{pages}{18750--18759}.
\newblock


\bibitem[Knoche et~al\mbox{.}(2023)]%
        {knoche2023explainable}
\bibfield{author}{\bibinfo{person}{Martin Knoche}, \bibinfo{person}{Torben Teepe}, \bibinfo{person}{Stefan H{\"o}rmann}, {and} \bibinfo{person}{Gerhard Rigoll}.} \bibinfo{year}{2023}\natexlab{}.
\newblock \showarticletitle{Explainable Model-Agnostic Similarity and Confidence in Face Verification}. In \bibinfo{booktitle}{\emph{Proceedings of the IEEE/CVF Winter Conference on Applications of Computer Vision}}. \bibinfo{pages}{711--718}.
\newblock


\bibitem[Kortylewski et~al\mbox{.}(2019)]%
        {kortylewski2019analyzing}
\bibfield{author}{\bibinfo{person}{Adam Kortylewski}, \bibinfo{person}{Bernhard Egger}, \bibinfo{person}{Andreas Schneider}, \bibinfo{person}{Thomas Gerig}, \bibinfo{person}{Andreas Morel-Forster}, {and} \bibinfo{person}{Thomas Vetter}.} \bibinfo{year}{2019}\natexlab{}.
\newblock \showarticletitle{Analyzing and reducing the damage of dataset bias to face recognition with synthetic data}. In \bibinfo{booktitle}{\emph{Proceedings of the IEEE/CVF Conference on Computer Vision and Pattern Recognition Workshops}}. \bibinfo{pages}{0--0}.
\newblock


\bibitem[Li et~al\mbox{.}(2018)]%
        {li2018tell}
\bibfield{author}{\bibinfo{person}{Kunpeng Li}, \bibinfo{person}{Ziyan Wu}, \bibinfo{person}{Kuan-Chuan Peng}, \bibinfo{person}{Jan Ernst}, {and} \bibinfo{person}{Yun Fu}.} \bibinfo{year}{2018}\natexlab{}.
\newblock \showarticletitle{Tell me where to look: Guided attention inference network}. In \bibinfo{booktitle}{\emph{Proceedings of the IEEE conference on computer vision and pattern recognition}}. \bibinfo{pages}{9215--9223}.
\newblock


\bibitem[Lin et~al\mbox{.}(2021)]%
        {lin2021xcos}
\bibfield{author}{\bibinfo{person}{Yu-Sheng Lin}, \bibinfo{person}{Zhe-Yu Liu}, \bibinfo{person}{Yu-An Chen}, \bibinfo{person}{Yu-Siang Wang}, \bibinfo{person}{Ya-Liang Chang}, {and} \bibinfo{person}{Winston~H Hsu}.} \bibinfo{year}{2021}\natexlab{}.
\newblock \showarticletitle{xCos: An explainable cosine metric for face verification task}.
\newblock \bibinfo{journal}{\emph{ACM Transactions on Multimedia Computing, Communications, and Applications (TOMM)}} \bibinfo{volume}{17}, \bibinfo{number}{3s} (\bibinfo{year}{2021}), \bibinfo{pages}{1--16}.
\newblock


\bibitem[Lu et~al\mbox{.}(2022)]%
        {lu2022novel}
\bibfield{author}{\bibinfo{person}{Yuhang Lu}, \bibinfo{person}{Luca Barras}, {and} \bibinfo{person}{Touradj Ebrahimi}.} \bibinfo{year}{2022}\natexlab{}.
\newblock \showarticletitle{A novel framework for assessment of deep face recognition systems in realistic conditions}. In \bibinfo{booktitle}{\emph{2022 10th European Workshop on Visual Information Processing (EUVIP)}}. IEEE, \bibinfo{pages}{1--6}.
\newblock


\bibitem[Lu et~al\mbox{.}(2024a)]%
        {lu2024explainable}
\bibfield{author}{\bibinfo{person}{Yuhang Lu}, \bibinfo{person}{Zewei Xu}, {and} \bibinfo{person}{Touradj Ebrahimi}.} \bibinfo{year}{2024}\natexlab{a}.
\newblock \showarticletitle{Explainable Face Verification via Feature-Guided Gradient Backpropagation}.
\newblock \bibinfo{journal}{\emph{arXiv preprint arXiv:2403.04549}} (\bibinfo{year}{2024}).
\newblock


\bibitem[Lu et~al\mbox{.}(2024b)]%
        {lu2024towards}
\bibfield{author}{\bibinfo{person}{Yuhang Lu}, \bibinfo{person}{Zewei Xu}, {and} \bibinfo{person}{Touradj Ebrahimi}.} \bibinfo{year}{2024}\natexlab{b}.
\newblock \showarticletitle{Towards visual saliency explanations of face verification}. In \bibinfo{booktitle}{\emph{Proceedings of the IEEE/CVF Winter Conference on Applications of Computer Vision}}. \bibinfo{pages}{4726--4735}.
\newblock


\bibitem[Maze et~al\mbox{.}(2018)]%
        {maze2018iarpa}
\bibfield{author}{\bibinfo{person}{Brianna Maze}, \bibinfo{person}{Jocelyn Adams}, \bibinfo{person}{James~A Duncan}, \bibinfo{person}{Nathan Kalka}, \bibinfo{person}{Tim Miller}, \bibinfo{person}{Charles Otto}, \bibinfo{person}{Anil~K Jain}, \bibinfo{person}{W~Tyler Niggel}, \bibinfo{person}{Janet Anderson}, \bibinfo{person}{Jordan Cheney}, {et~al\mbox{.}}} \bibinfo{year}{2018}\natexlab{}.
\newblock \showarticletitle{Iarpa janus benchmark-c: Face dataset and protocol}. In \bibinfo{booktitle}{\emph{2018 international conference on biometrics (ICB)}}. IEEE, \bibinfo{pages}{158--165}.
\newblock


\bibitem[Meng et~al\mbox{.}(2021)]%
        {meng2021magface}
\bibfield{author}{\bibinfo{person}{Qiang Meng}, \bibinfo{person}{Shichao Zhao}, \bibinfo{person}{Zhida Huang}, {and} \bibinfo{person}{Feng Zhou}.} \bibinfo{year}{2021}\natexlab{}.
\newblock \showarticletitle{Magface: A universal representation for face recognition and quality assessment}. In \bibinfo{booktitle}{\emph{Proceedings of the IEEE/CVF Conference on Computer Vision and Pattern Recognition}}. \bibinfo{pages}{14225--14234}.
\newblock


\bibitem[Mery(2022)]%
        {mery2022true}
\bibfield{author}{\bibinfo{person}{Domingo Mery}.} \bibinfo{year}{2022}\natexlab{}.
\newblock \showarticletitle{True black-box explanation in facial analysis}. In \bibinfo{booktitle}{\emph{Proceedings of the IEEE/CVF Conference on Computer Vision and Pattern Recognition}}. \bibinfo{pages}{1596--1605}.
\newblock


\bibitem[Mery and Morris(2022)]%
        {mery2022black}
\bibfield{author}{\bibinfo{person}{Domingo Mery} {and} \bibinfo{person}{Bernardita Morris}.} \bibinfo{year}{2022}\natexlab{}.
\newblock \showarticletitle{On black-box explanation for face verification}. In \bibinfo{booktitle}{\emph{Proceedings of the IEEE/CVF Winter Conference on Applications of Computer Vision}}. \bibinfo{pages}{3418--3427}.
\newblock


\bibitem[Olah et~al\mbox{.}(2017)]%
        {olah2017feature}
\bibfield{author}{\bibinfo{person}{Chris Olah}, \bibinfo{person}{Alexander Mordvintsev}, {and} \bibinfo{person}{Ludwig Schubert}.} \bibinfo{year}{2017}\natexlab{}.
\newblock \showarticletitle{Feature Visualization}.
\newblock \bibinfo{journal}{\emph{Distill}} (\bibinfo{year}{2017}).
\newblock
\urldef\tempurl%
\url{https://doi.org/10.23915/distill.00007}
\showDOI{\tempurl}
\newblock
\shownote{https://distill.pub/2017/feature-visualization}.


\bibitem[Petsiuk et~al\mbox{.}(2018)]%
        {petsiuk2018rise}
\bibfield{author}{\bibinfo{person}{Vitali Petsiuk}, \bibinfo{person}{Abir Das}, {and} \bibinfo{person}{Kate Saenko}.} \bibinfo{year}{2018}\natexlab{}.
\newblock \showarticletitle{{RISE:} Randomized Input Sampling for Explanation of Black-box Models}. In \bibinfo{booktitle}{\emph{British Machine Vision Conference 2018, {BMVC} 2018, Newcastle, UK, September 3-6, 2018}}. \bibinfo{pages}{151}.
\newblock


\bibitem[Petsiuk et~al\mbox{.}(2021)]%
        {petsiuk2021black}
\bibfield{author}{\bibinfo{person}{Vitali Petsiuk}, \bibinfo{person}{Rajiv Jain}, \bibinfo{person}{Varun Manjunatha}, \bibinfo{person}{Vlad~I Morariu}, \bibinfo{person}{Ashutosh Mehra}, \bibinfo{person}{Vicente Ordonez}, {and} \bibinfo{person}{Kate Saenko}.} \bibinfo{year}{2021}\natexlab{}.
\newblock \showarticletitle{Black-box explanation of object detectors via saliency maps}. In \bibinfo{booktitle}{\emph{Proceedings of the IEEE/CVF Conference on Computer Vision and Pattern Recognition}}. \bibinfo{pages}{11443--11452}.
\newblock


\bibitem[Ribeiro et~al\mbox{.}(2016)]%
        {ribeiro2016should}
\bibfield{author}{\bibinfo{person}{Marco~Tulio Ribeiro}, \bibinfo{person}{Sameer Singh}, {and} \bibinfo{person}{Carlos Guestrin}.} \bibinfo{year}{2016}\natexlab{}.
\newblock \showarticletitle{" Why should i trust you?" Explaining the predictions of any classifier}. In \bibinfo{booktitle}{\emph{Proceedings of the 22nd ACM SIGKDD international conference on knowledge discovery and data mining}}. \bibinfo{pages}{1135--1144}.
\newblock


\bibitem[Selvaraju et~al\mbox{.}(2017)]%
        {selvaraju2017grad}
\bibfield{author}{\bibinfo{person}{Ramprasaath~R Selvaraju}, \bibinfo{person}{Michael Cogswell}, \bibinfo{person}{Abhishek Das}, \bibinfo{person}{Ramakrishna Vedantam}, \bibinfo{person}{Devi Parikh}, {and} \bibinfo{person}{Dhruv Batra}.} \bibinfo{year}{2017}\natexlab{}.
\newblock \showarticletitle{Grad-cam: Visual explanations from deep networks via gradient-based localization}. In \bibinfo{booktitle}{\emph{Proceedings of the IEEE international conference on computer vision}}. \bibinfo{pages}{618--626}.
\newblock


\bibitem[Selvaraju et~al\mbox{.}(2016)]%
        {selvaraju2016grad}
\bibfield{author}{\bibinfo{person}{Ramprasaath~R Selvaraju}, \bibinfo{person}{Abhishek Das}, \bibinfo{person}{Ramakrishna Vedantam}, \bibinfo{person}{Michael Cogswell}, \bibinfo{person}{Devi Parikh}, {and} \bibinfo{person}{Dhruv Batra}.} \bibinfo{year}{2016}\natexlab{}.
\newblock \showarticletitle{Grad-CAM: Why did you say that?}
\newblock \bibinfo{journal}{\emph{arXiv preprint arXiv:1611.07450}} (\bibinfo{year}{2016}).
\newblock


\bibitem[Simonyan et~al\mbox{.}(2013)]%
        {simonyan2013deep}
\bibfield{author}{\bibinfo{person}{Karen Simonyan}, \bibinfo{person}{Andrea Vedaldi}, {and} \bibinfo{person}{Andrew Zisserman}.} \bibinfo{year}{2013}\natexlab{}.
\newblock \showarticletitle{Deep inside convolutional networks: Visualising image classification models and saliency maps}.
\newblock \bibinfo{journal}{\emph{arXiv preprint arXiv:1312.6034}} (\bibinfo{year}{2013}).
\newblock


\bibitem[Stylianou et~al\mbox{.}(2019)]%
        {stylianou2019visualizing}
\bibfield{author}{\bibinfo{person}{Abby Stylianou}, \bibinfo{person}{Richard Souvenir}, {and} \bibinfo{person}{Robert Pless}.} \bibinfo{year}{2019}\natexlab{}.
\newblock \showarticletitle{Visualizing deep similarity networks}. In \bibinfo{booktitle}{\emph{2019 IEEE winter conference on applications of computer vision (WACV)}}. IEEE, \bibinfo{pages}{2029--2037}.
\newblock


\bibitem[Terh{\"o}rst et~al\mbox{.}(2021)]%
        {terhorst2021comprehensive}
\bibfield{author}{\bibinfo{person}{Philipp Terh{\"o}rst}, \bibinfo{person}{Jan~Niklas Kolf}, \bibinfo{person}{Marco Huber}, \bibinfo{person}{Florian Kirchbuchner}, \bibinfo{person}{Naser Damer}, \bibinfo{person}{Aythami~Morales Moreno}, \bibinfo{person}{Julian Fierrez}, {and} \bibinfo{person}{Arjan Kuijper}.} \bibinfo{year}{2021}\natexlab{}.
\newblock \showarticletitle{A comprehensive study on face recognition biases beyond demographics}.
\newblock \bibinfo{journal}{\emph{IEEE Transactions on Technology and Society}} \bibinfo{volume}{3}, \bibinfo{number}{1} (\bibinfo{year}{2021}), \bibinfo{pages}{16--30}.
\newblock


\bibitem[Wang and Deng(2021)]%
        {wang2021deep}
\bibfield{author}{\bibinfo{person}{Mei Wang} {and} \bibinfo{person}{Weihong Deng}.} \bibinfo{year}{2021}\natexlab{}.
\newblock \showarticletitle{Deep face recognition: A survey}.
\newblock \bibinfo{journal}{\emph{Neurocomputing}}  \bibinfo{volume}{429} (\bibinfo{year}{2021}), \bibinfo{pages}{215--244}.
\newblock


\bibitem[Williford et~al\mbox{.}(2020)]%
        {williford2020explainable}
\bibfield{author}{\bibinfo{person}{Jonathan~R Williford}, \bibinfo{person}{Brandon~B May}, {and} \bibinfo{person}{Jeffrey Byrne}.} \bibinfo{year}{2020}\natexlab{}.
\newblock \showarticletitle{Explainable face recognition}. In \bibinfo{booktitle}{\emph{Computer Vision--ECCV 2020: 16th European Conference, Glasgow, UK, August 23--28, 2020, Proceedings, Part XI}}. Springer, \bibinfo{pages}{248--263}.
\newblock


\bibitem[Xu et~al\mbox{.}(2023)]%
        {xu2023discriminative}
\bibfield{author}{\bibinfo{person}{Zewei Xu}, \bibinfo{person}{Yuhang Lu}, {and} \bibinfo{person}{Touradj Ebrahimi}.} \bibinfo{year}{2023}\natexlab{}.
\newblock \showarticletitle{Discriminative Deep Feature Visualization for Explainable Face Recognition}. In \bibinfo{booktitle}{\emph{25th {IEEE} International Workshop on Multimedia Signal Processing, {MMSP} 2023, Poitiers, France, September 27-29, 2023}}. \bibinfo{pages}{1--6}.
\newblock


\bibitem[Yin et~al\mbox{.}(2019)]%
        {yin2019towards}
\bibfield{author}{\bibinfo{person}{Bangjie Yin}, \bibinfo{person}{Luan Tran}, \bibinfo{person}{Haoxiang Li}, \bibinfo{person}{Xiaohui Shen}, {and} \bibinfo{person}{Xiaoming Liu}.} \bibinfo{year}{2019}\natexlab{}.
\newblock \showarticletitle{Towards interpretable face recognition}. In \bibinfo{booktitle}{\emph{Proceedings of the IEEE/CVF International Conference on Computer Vision}}. \bibinfo{pages}{9348--9357}.
\newblock


\bibitem[Zeiler and Fergus(2014)]%
        {zeiler2014visualizing}
\bibfield{author}{\bibinfo{person}{Matthew~D Zeiler} {and} \bibinfo{person}{Rob Fergus}.} \bibinfo{year}{2014}\natexlab{}.
\newblock \showarticletitle{Visualizing and understanding convolutional networks}. In \bibinfo{booktitle}{\emph{Computer Vision--ECCV 2014: 13th European Conference, Zurich, Switzerland, September 6-12, 2014, Proceedings, Part I 13}}. Springer, \bibinfo{pages}{818--833}.
\newblock


\bibitem[Zhang et~al\mbox{.}(2018)]%
        {zhang2018top}
\bibfield{author}{\bibinfo{person}{Jianming Zhang}, \bibinfo{person}{Sarah~Adel Bargal}, \bibinfo{person}{Zhe Lin}, \bibinfo{person}{Jonathan Brandt}, \bibinfo{person}{Xiaohui Shen}, {and} \bibinfo{person}{Stan Sclaroff}.} \bibinfo{year}{2018}\natexlab{}.
\newblock \showarticletitle{Top-down neural attention by excitation backprop}.
\newblock \bibinfo{journal}{\emph{International Journal of Computer Vision}} \bibinfo{volume}{126}, \bibinfo{number}{10} (\bibinfo{year}{2018}), \bibinfo{pages}{1084--1102}.
\newblock


\bibitem[Zheng and Deng(2018)]%
        {zheng2018cross}
\bibfield{author}{\bibinfo{person}{Tianyue Zheng} {and} \bibinfo{person}{Weihong Deng}.} \bibinfo{year}{2018}\natexlab{}.
\newblock \showarticletitle{Cross-pose lfw: A database for studying cross-pose face recognition in unconstrained environments}.
\newblock \bibinfo{journal}{\emph{Beijing University of Posts and Telecommunications, Tech. Rep}}  \bibinfo{volume}{5} (\bibinfo{year}{2018}), \bibinfo{pages}{7}.
\newblock


\bibitem[Zhou et~al\mbox{.}(2016)]%
        {zhou2016learning}
\bibfield{author}{\bibinfo{person}{Bolei Zhou}, \bibinfo{person}{Aditya Khosla}, \bibinfo{person}{Agata Lapedriza}, \bibinfo{person}{Aude Oliva}, {and} \bibinfo{person}{Antonio Torralba}.} \bibinfo{year}{2016}\natexlab{}.
\newblock \showarticletitle{Learning deep features for discriminative localization}. In \bibinfo{booktitle}{\emph{Proceedings of the IEEE conference on computer vision and pattern recognition}}. \bibinfo{pages}{2921--2929}.
\newblock


\bibitem[Zhuang et~al\mbox{.}(2010)]%
        {zhuang2010facial}
\bibfield{author}{\bibinfo{person}{Ziqing Zhuang}, \bibinfo{person}{Douglas Landsittel}, \bibinfo{person}{Stacey Benson}, \bibinfo{person}{Raymond Roberge}, {and} \bibinfo{person}{Ronald Shaffer}.} \bibinfo{year}{2010}\natexlab{}.
\newblock \showarticletitle{Facial anthropometric differences among gender, ethnicity, and age groups}.
\newblock \bibinfo{journal}{\emph{Annals of occupational hygiene}} \bibinfo{volume}{54}, \bibinfo{number}{4} (\bibinfo{year}{2010}), \bibinfo{pages}{391--402}.
\newblock


\end{thebibliography}










\end{document}